\documentclass[journal]{IEEEtran}
\usepackage{graphicx}
\usepackage{amsmath,amssymb,amsfonts}
\usepackage{booktabs}
\usepackage{array}
\usepackage{tabularx}
\usepackage{xcolor}
\usepackage{url}
\usepackage{float}
\usepackage{placeins}
\usepackage{microtype}
\usepackage[colorlinks=false,hidelinks]{hyperref}
\graphicspath{{./}}
\newcommand{\methodfull}{Underwater C\textsuperscript{3}-JEPA}
\newcommand{\method}{C\textsuperscript{3}-JEPA}
\newcommand{\gocvf}{\textsc{GO-CVF}}
\newcommand{\ocpd}{\textsc{OCPD}}

\begin{document}
\title{\methodfull{}: An Object-Centric Cross-View World Model for ROV Salvage}
\hypersetup{pdftitle={Underwater C3-JEPA: An Object-Centric Cross-View World Model for ROV Salvage},pdfauthor={Yuncong Yang, Jinlong Li, Yulong Xue, Feng Wu, Chunwen Zhang, Lei Qiao, Xuyang Wang}}
\author{Yuncong~Yang, Jinlong~Li, Yulong~Xue, Feng~Wu, Chunwen~Zhang, Lei~Qiao, and Xuyang~Wang%
\thanks{Yuncong Yang, Jinlong Li, Yulong Xue, Feng Wu, Chunwen Zhang, Lei Qiao, and Xuyang Wang are with the Underwater Engineering Institute, Shanghai Jiao Tong University, Shanghai 200240, China (e-mail: qiaolei@sjtu.edu.cn; wangxuyang@sjtu.edu.cn).}%
\thanks{Jinlong Li and Yulong Xue contributed equally to this work and share the second authorship; Feng Wu and Chunwen Zhang contributed equally to this work and share the third authorship. (Corresponding authors: Lei Qiao and Xuyang Wang.)}%
}
\maketitle
\begingroup
\renewcommand\thefootnote{}
\footnote{This work has been submitted to the IEEE for possible publication. Copyright may be transferred without notice, after which this version may no longer be accessible.}
\endgroup
\begin{abstract}
We present \methodfull{} (cross-view, control-conditioned, context-extended), an object-centric multi-view predictive world model for near-field heavy-load underwater ROV salvage. Without contact sensors, it predicts in latent space how the task-object state evolves through contact interaction and under the hydrodynamic lag of the vehicle, from synchronized multi-view RGB observations and vehicle control signals. \method{} encodes multi-camera observations into task-object and context tokens, fuses cross-camera evidence through held-out-view attention, and directly predicts future states conditioned on control. Weak binding anchors the target and gripper at low annotation cost, while SIGReg sharpens the geometric representation. Experiments show that the learned representation transfers substantially more task-relevant information to downstream probes than a reconstruction-free latent baseline, while keeping the predictor lightweight. The resulting predictive interface supports model-predictive-control (MPC) candidate evaluation and imagined-rollout behavior-agent training. Validation on real underwater video shows the same architecture recovering a withheld camera's object state and staying ahead of persistence, so the recipe transfers beyond simulation.

\textbf{Note to Practitioners}---Recovering a large sunken structure with a work-class ROV is decided in the final meters of approach, where the target is only partially visible and the vehicle answers each thrust command with a visible delay. This paper develops a predictive model for that setting. Rather than reconstructing the scene, it tracks the two task objects that matter, target and gripper, in a compact representation pooled across the vehicle's cameras, learning directly from ordinary mission recordings how that representation evolves under the pilot's commands. Because the model is small, it can run in real time and score candidate control sequences before they are executed, which is the capability a conventional autopilot lacks. Three practical consequences follow: no new sensors are required, since training uses the video and telemetry an ROV already logs; losing the target in one camera is tolerated because the other cameras' evidence is combined; and the recipe transfers to real footage with the simulator out of the loop. The limitations are straightforward: the field trials were open-loop, underwater accuracy is expressed in relative units, and closed-loop success rates at sea have yet to be measured. The same interface should extend to other near-contact subsea tasks---inspection, connector mating, tool deployment---whenever a work-class ROV must act on a partially visible target.
\end{abstract}

\section{Introduction}\label{sec:intro}

\begin{figure*}[t!]
\centering
\begin{tabular}{@{}ccc@{}}
\includegraphics[height=4.3cm]{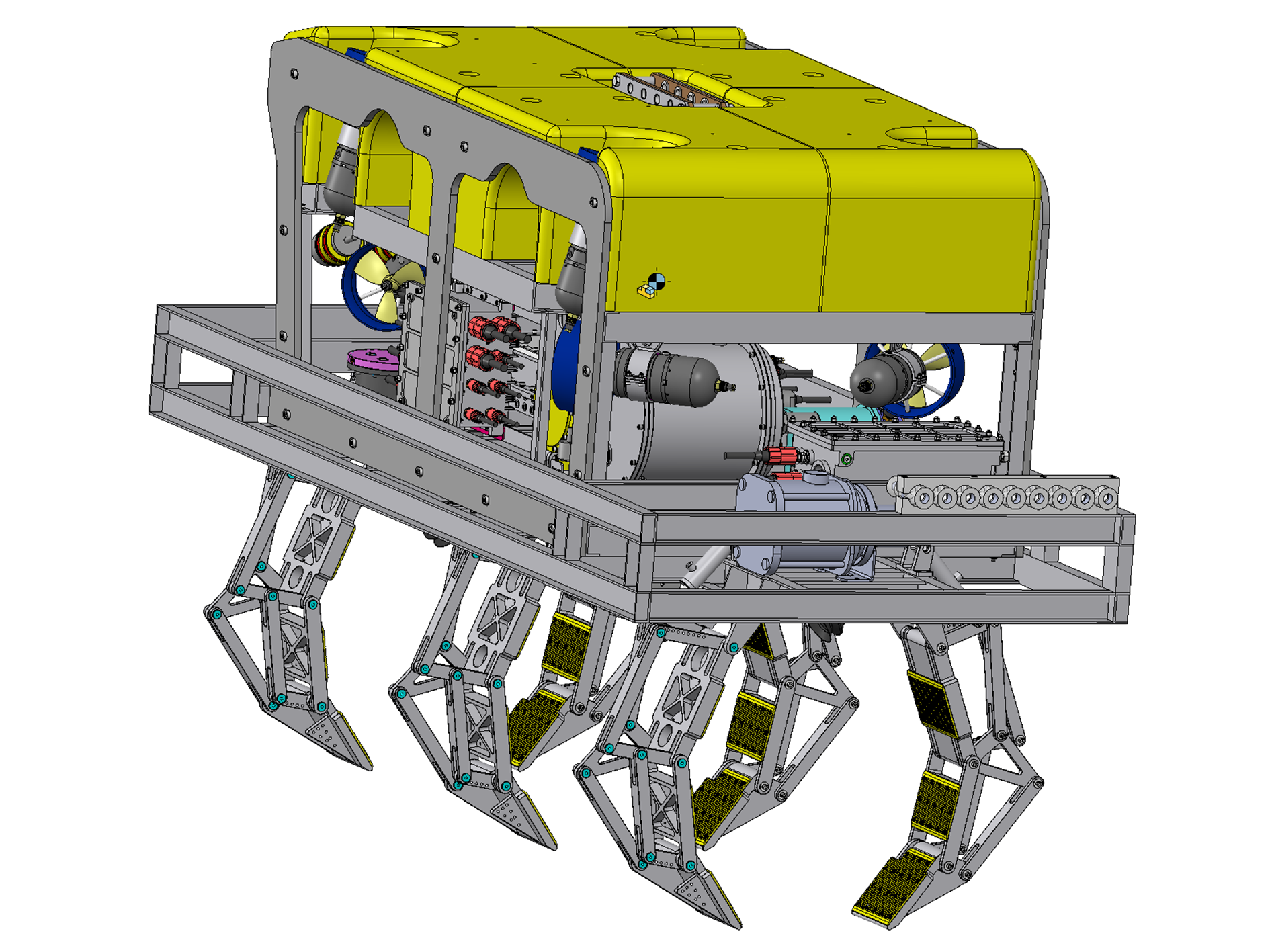} & \includegraphics[height=4.3cm]{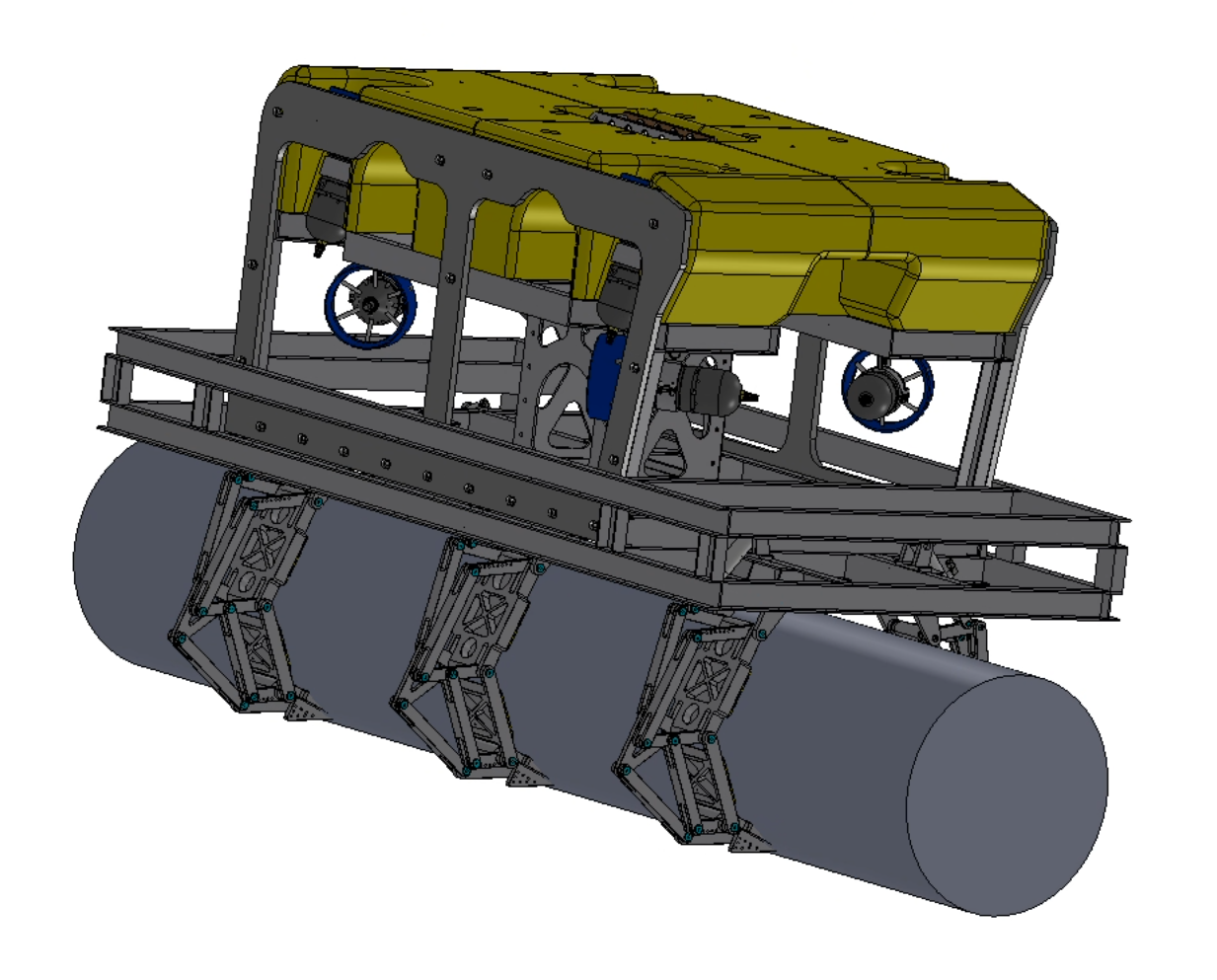} & \includegraphics[height=4.3cm]{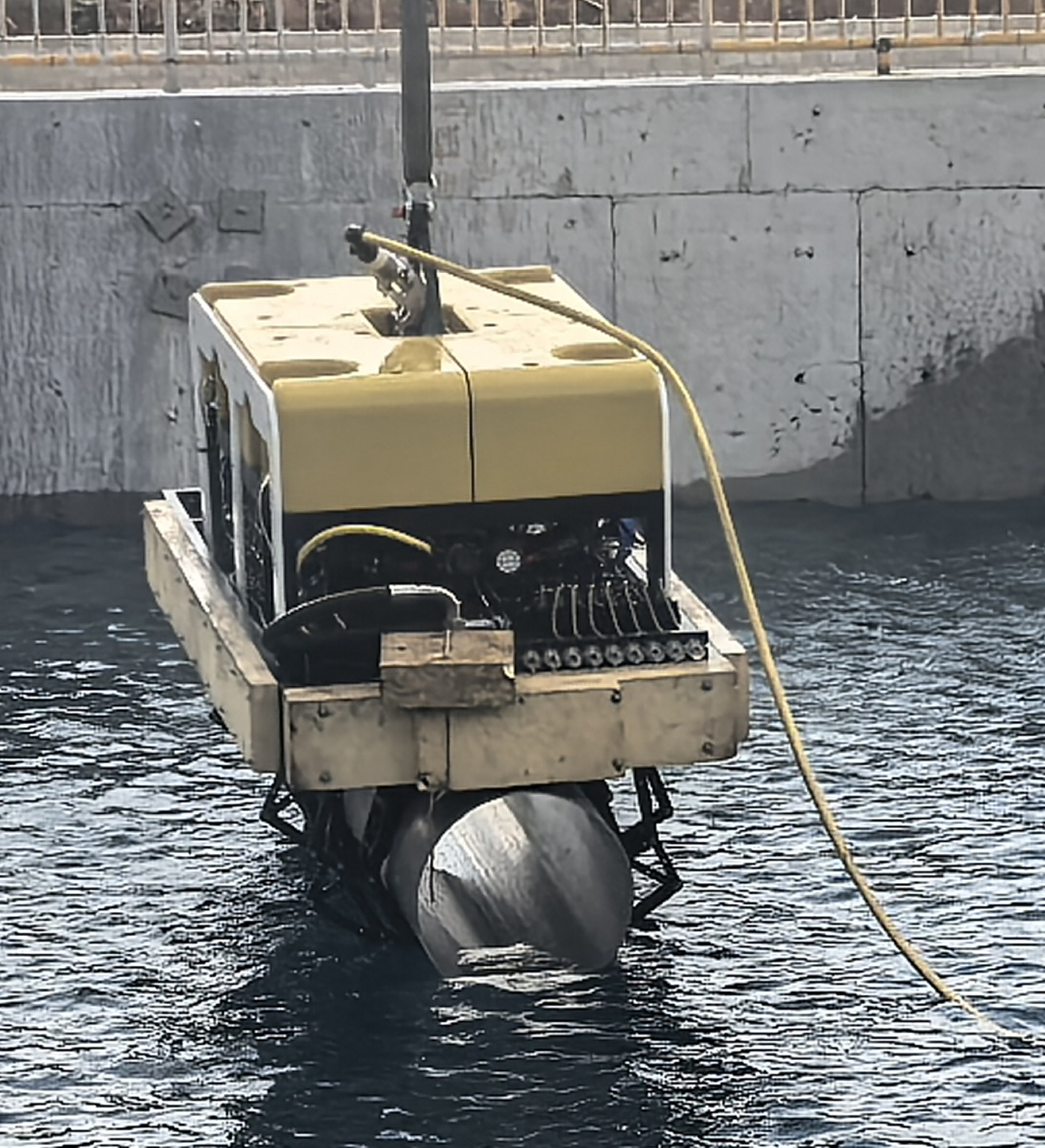}\\[1pt]
{\small (a)} & {\small (b)} & {\small (c)}
\end{tabular}
\caption{The heavy-duty underwater salvage ROV used in this study: (a) isometric view; (b) grasping a target object (CAD rendering); (c) lifting a target object in a real-environment trial.}
\label{fig:rov_platform}
\end{figure*}

Deep-sea salvage of large sunken objects is time-critical: contamination grows while a wreck lies on the seabed \cite{Cong_2021}. Industrial practice rigs slings with a work-class ROV and lifts with a crane---slow, intervention-heavy, damaging to a fragile structure \cite{Simetti_2021,Sun_2024,Morgan_2022}; carrying the gripper on the vehicle removes that stage but forces the ROV to envelope a large, shape-uncertain object from a hydrodynamically lagged base. Our platform is atypical: not a dexterous arm but paired linkage-driven underactuated grippers closing downward, trading degrees of freedom for load capacity \cite{Birglen_2008,Bicchi_2000,Yang_2025_Salvage}, with coupled, contact-dependent, partly closed-loop kinematics and no contact sensing. Fig.~\ref{fig:rov_platform} shows the platform and its gripper. The mission is therefore staged for this platform: acquisition, approach, alignment, enveloping contact, lift/hold, and verification, all specific to this vehicle--tool--target combination.

In the decisive near-field window the target can sit at any bearing across the cameras' combined field of view, occluded by hull and gripper and only partially visible in any single view under turbidity and low contrast \cite{Cong_2021,Huy_2023}. Mapping first and perceiving in a point cloud fits poorly: visual SLAM drifts, loses tracking under backscatter, and needs loop closures a short approach never revisits \cite{Campos_2021_ORBSLAM3,Zhang_2022_UVSSLAM,Joshi_2019_VIOCompare,Rahman_2022_SVIn2}; structured light is denser but demands short standoff, a calibrated projector, and dwell-and-scan motion \cite{Lyu_2023_StructuredLight,Ou_2023_WaterMBSL}. Both approaches reconstruct more than the decision requires.

Control is constrained differently. Real-system learning must succeed from few trials, at control rate, under large but unknown actuation delays \cite{DulacArnold_2019,Chua_2018_PETS}, and simulation handles this gripper poorly: closed kinematic loops stiffen the constrained dynamics into tiny steps, soft-constraint relaxation, or outright solver failure \cite{Featherstone_2008,Todorov_2012_MuJoCo}. Tactile instrumentation is fragile and costly, and needs pressure housing and cabling on a jaw closing on rough wreckage \cite{Yuan_2017_GelSight,Lambeta_2020_DIGIT}. Neither route internalizes this behavior affordably.

\methodfull{} changes what must be reconstructed, building an \emph{object-centric perception domain} (\ocpd{}): task-object classes (target and gripper) are held at full fidelity and forcibly aligned across cameras, while remaining evidence stays in free context slots kept alongside without alignment. We call this routing guided object-centric cross-view fusion (\gocvf{}). Because the decomposition fixes which entities receive tokens, fusion fits an ROV budget and new task classes are synthesized from the same weak-anchor route rather than exhaustively reconstructed; anchors are DINOv3-clustering proposals, with human review reserved for ambiguous frames.

A second element provides the physical dynamics in two roles: offline, the control-conditioned world model synthesizes imagined rollouts so a behavior agent improves without extra ship time \cite{Hafner_2020_Dreamer,Hafner_2023,Wu_2023_DayDreamer}; online, it ranks candidate commands for multi-trajectory reasoning. One model serves both: driven by commands rather than by measured motion, the predictor learns the thrust-buildup and added-mass lag that short-horizon models leave unmodeled \cite{Fossen_2021}; predicting fused tokens instead of pixels keeps it at $1.33$\,M parameters.

Sonar, navigation and force sensing all matter; this paper asks the narrower question: whether the on-board visual channels compress into a state that keeps the target and gripper identifiable across views and through contact and supports rollouts.

Our contributions are:
\begin{enumerate}
    \item \textbf{Object-centric perception domain (\ocpd{}) and guided object-centric cross-view fusion (\gocvf{}):} weak anchors stabilize identity, held-out-view attention consolidates partial evidence, and free slots keep unaligned context alongside; because alignment is spent only where it pays, fusion fits an ROV budget and new task classes are synthesized cheaply.
    \item \textbf{A control-conditioned JEPA world model of the vehicle--object dynamics, with a model-predictive-control (MPC) interface:} driving the predictor by commands makes the actuation lag between command and realized motion part of the learned state; a $1.33$\,M footprint lets one model serve imagined training on limited offline data and candidate-command ranking for model-predictive control; and each ingredient of the loop---predictor, latent geometry and ranking signal---is validated separately.
    \item \textbf{Validation on real underwater video:} two field deployments on uncalibrated cameras confirm that the hidden-view recovery and the control-conditioned rollout carry beyond simulation.
\end{enumerate}

\section{Related Work}

\subsection{Underwater perception: scene mapping versus task-object state}
Underwater perception for intervention is conventionally organized around mapping. Visual, visual-inertial, and acoustic-inertial SLAM systems estimate vehicle pose and build sparse or semi-dense structure, with underwater variants adding image enhancement, pressure-depth terms, and sonar or DVL factors to counter drift and tracking loss \cite{Campos_2021_ORBSLAM3,Joshi_2019_VIOCompare,Rahman_2022_SVIn2,Zhang_2022_UVSSLAM}; where dense near-field geometry is required, active projection and structured-light scanners deliver higher per-frame accuracy at the cost of short standoff, sensitivity to scattering and turbidity, and usually a dwell-or-scan motion constraint \cite{Lyu_2023_StructuredLight,Ou_2023_WaterMBSL}.  Both families reconstruct the scene first and interpret it afterwards. \method{} takes the complementary route and never commits to a metric map: it maintains an object-centric perception domain (\ocpd{}) in which only task objects are resolved and aligned. The two answers are not adversarial---mapping remains right for search and survey---but differ in cost and in what survives occlusion and contact.

\subsection{Object-centric visual representation}
Slot Attention decomposes a scene into a small unordered set of entity-like vectors \cite{Locatello_2020}. SAVi and SAVi++ extend object-centric learning through time \cite{Kipf_2021_SAVi,Elsayed_2022_SAVipp}; DINOSAUR and VideoSAUR show that reconstructing self-supervised features can support object discovery in natural images and videos \cite{Seitzer_2023_DINOSAUR,Zadaianchuk_2023}; and SlotFormer studies long-range dynamics in object slots \cite{Wu_2023_SlotFormer}. A practical difficulty for robotics is that unconstrained slots may permute or absorb a small task object. We use weak binding only for the target and gripper (a single extra binding-only slot handles the fixed ROV frame), leaving the remaining slots to account for scene context. This is deliberately lighter than dense segmentation: DINOv3-feature clustering proposes masks from a few region prompts. An orthogonal upgrade path replaces instance slots with task-functional roles for planning \cite{Cheng_2026_SRWm}; our design keeps two instance-level roles (target and gripper), sufficient for enveloping salvage, with the mechanism validated on real frames.

\subsection{Multi-view fusion and underwater predictive models}
Multi-view object-centric models such as ROOTS and MulMON aggregate partial observations and test unobserved viewpoints \cite{Chen_2020_ROOTS,Li_2021_MulMON}, and the held-out-view principle is well established at the pixel level: cross-view completion predicts the masked content of one image from a second view of the same scene \cite{Weinzaepfel_2022_CroCo}, and multi-view masked autoencoders reconstruct withheld viewpoints before a world model is trained on the resulting representations \cite{Seo_2023_MVMWM}. We adapt the idea to cameras rigidly mounted on one moving ROV: instead of forcing a shared token to decode identically in all cameras, the target view is withheld from cross-view keys and values, and a view-conditioned decoder reconstructs its features. This discourages direct copying while preserving legitimate view-dependent appearance. AquaJEPA combines RGB, forward sonar and proprioception for simulated robot dynamics \cite{Gazzaev_2026_AquaJEPA}, a setting outside the current \method{} data interface. A recent AUV docking study combines hand-built hydrodynamic priors with a compact learned model under communication constraints \cite{Zhang_2026_PICWM}; our route is fully learned but shares the goal of onboard-scale predictive control.

\subsection{Latent prediction and planning}
A world model is a functional role rather than a single architecture: a learned state, explicit or latent, whose evolution can be queried under candidate actions. It is used for several related purposes: compressing high-dimensional observations into a state for control, predicting short-horizon consequences for model-predictive control (MPC), evaluating counterfactual action sequences, training a behavior policy in imagined rollouts, and transferring a reusable state representation to downstream estimators. Designs differ along three overlapping axes: pixel-space generative models that reconstruct future observations; latent dynamics models that predict a compact embedding; and object-centric models whose state is partitioned into entities, relations, and context. 

Early latent-dynamics systems such as PlaNet learn a compact transition model from pixels for planning, while Dreamer extends the idea to behavior learning through latent imagination \cite{Hafner_2019_PlaNet,Hafner_2020_Dreamer,Hafner_2023}. Visual Foresight uses action-conditioned visual prediction to select robot motions, and TD-MPC2 repeatedly optimizes short candidate sequences in a learned latent space \cite{Finn_2017_VisualForesight,Ebert_2018_VisualMPC,Hansen_2024}. These approaches establish the broad world-model planning loop---observe, predict candidate consequences, score them, execute a short prefix, re-observe---but do not by themselves guarantee that the latent state preserves target and gripper identity across occlusion and contact. For underwater manipulation this matters: a representation can predict average motion while discarding the pose, joint, or object evidence a downstream controller needs.

JEPA methods shift the prediction target from pixels to representations. I-JEPA predicts masked image embeddings without reconstructing every pixel, and V-JEPA~2 extends this principle to video understanding, action conditioning, and planning \cite{Assran_2023_IJEPA,Assran_2025_VJEPA2}. LeJEPA frames joint-embedding prediction as a geometrically regularized objective, while AdaJEPA studies adaptive latent prediction; both motivate treating the latent geometry as a first-class design object \cite{Balestriero_2025,Wang_2026_AdaJEPA}. The efficiency is attractive for an ROV decision interface---the predictor stays small while candidates are rolled out in state space---but a low latent prediction error does not imply object semantics, observable calibration, or task-relevant information transfer. The closest published analogue is SkyJEPA, which maps frozen JEPA latents to interpretable state with a physics-inspired prober and carries it to real quadrotor control without fine-tuning \cite{Rao_2026_SkyJEPA}; our frozen geometry head plays the analogous role for salvage.

Causal-JEPA makes the control question more explicit by studying object-level latent interventions, asking how changes in an object or intervention variable alter the predicted future \cite{Nam_2026}. It builds on object-centric encoders such as VideoSAUR or SAVi; we take the VideoSAUR branch because it yields slots in real-world video without the motion or depth conditioning SAVi-style pipelines assume \cite{Seitzer_2023_DINOSAUR,Zadaianchuk_2023}. This work combines the three ingredients: weak binding preserves target and gripper identity, held-out-view fusion preserves complementary evidence, and the latent predictor exposes candidate consequences to MPC and imagined-rollout evaluation.

\begin{figure*}[t!]
\centering
\includegraphics[width=0.76\textwidth]{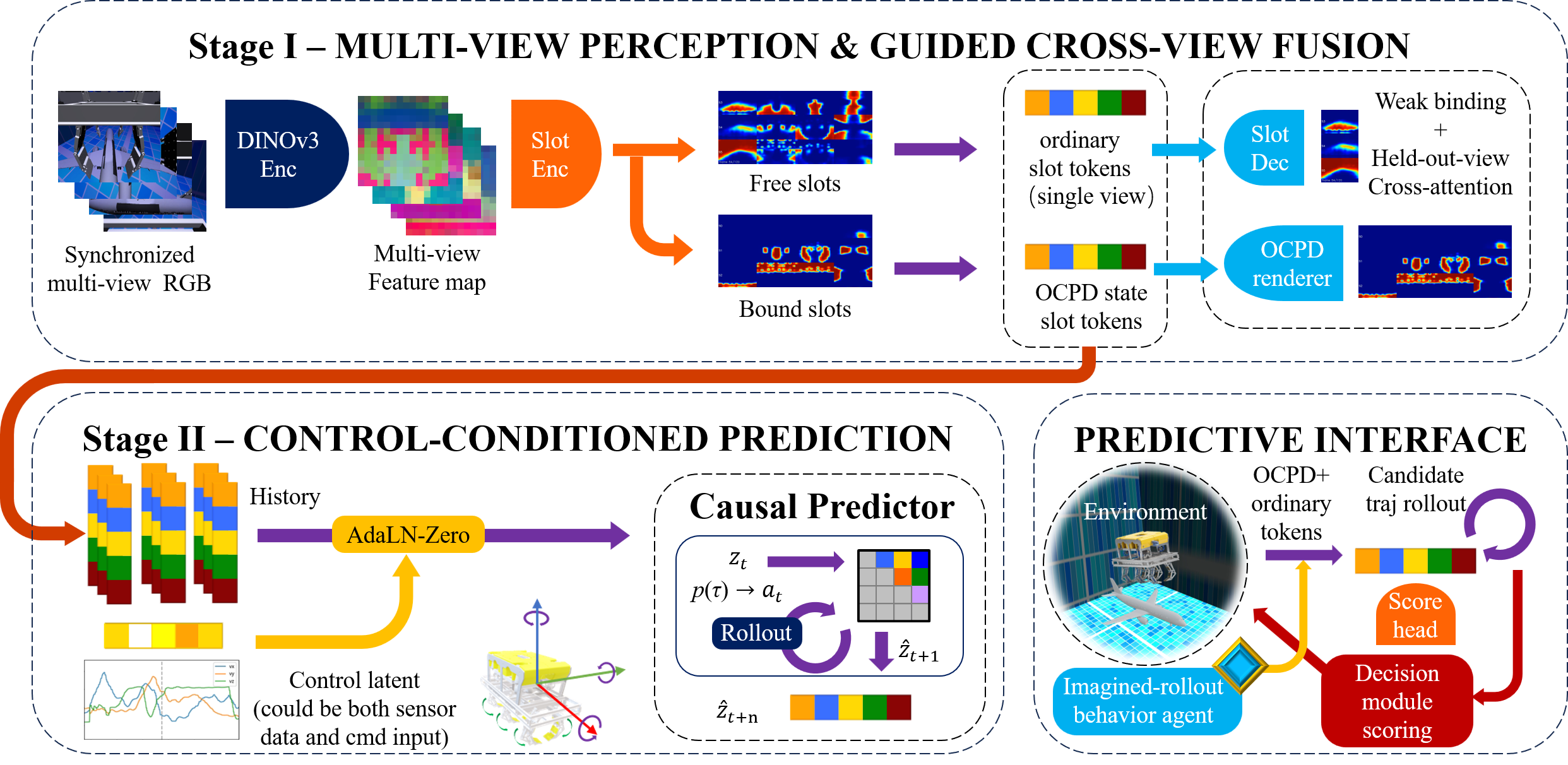}
\caption{\method{} architecture overview. Stage~I maps synchronized multi-view RGB to bound task-object slots and free context slots; Stage~II injects the control latent through AdaLN-Zero and predicts the state autoregressively. Right: the predictive interface.}
\label{fig:architecture}
\end{figure*}

\noindent\textbf{Reconstruction-free latent world models.} LeWM-style architectures train a JEPA encoder--predictor from raw pixels with a next-embedding prediction loss and a SIGReg prior, without any reconstruction or object-centric structure \cite{Maes_2026_LeWM}. We train such a model from scratch on exactly the same multi-view recordings and control signals as \method{}, and compare both representations under identical downstream probes. This head-to-head comparison, reported in Sec.~\ref{sec:lewm_compare}, isolates what reconstruction-guided, object-centric supervision adds over purely latent regularization at this scale. Training-time geometric priors (paired-depth alignment) similarly improve JEPA representations from real robot data while leaving inference RGB-only \cite{Khan_2026_DepthReg}. In our recipe, geometry enters through the frozen reconstruction decoder instead of an auxiliary loss.

\section{Method}

\subsection{Architecture overview}
Fig.~\ref{fig:architecture} summarizes the pipeline. At time $t$, synchronized camera observations are $I_t^{1:V}$. A frozen DINOv3-L feature extractor \cite{Simeoni_2025} maps each view to patch features $X_t^v \in \mathbb{R}^{P\times d_p}$, and a temporal slot encoder returns $K$ slots $S_t^v \in \mathbb{R}^{K\times d_s}$. In simulation $V=4$; the architecture itself accepts the subset of views available on a platform. Three binding slots represent the target UUV, the gripper, and the fixed ROV frame. Because the frame occludes parts of every view at a fixed pose, it is kept in a binding-only slot (slot~2), excluded from cross-view fusion and from the predictive state; the free slots are not spent on it. Cross-view attention produces shared target and gripper tokens, and the predictive state concatenates these with the three context slots of a single reference view, $v_r$ (the rear camera, which also serves as the free-slot prediction target),
\begin{equation}
z_t=[\bar{s}_{t}^{\mathrm{target}},\bar{s}_{t}^{\mathrm{grip}},s_t^{v_r,3},s_t^{v_r,4},s_t^{v_r,5}].
\label{eq:ocpd_state}
\end{equation}
Equation~\eqref{eq:ocpd_state} is the object-centric perception domain (\ocpd{}) introduced in Sec.~\ref{sec:intro}: the two task-object entries carry the target and gripper at full fidelity and are forcibly aligned across cameras, while the three context entries are free slots carried without alignment. The state is object-centric by construction rather than a uniformly compressed scene descriptor, which keeps cross-view fusion inside an ROV compute budget.

\subsection{Object-centric slots and geometric regularization}
\begin{figure*}[t!]
\centering
\includegraphics[width=0.76\textwidth]{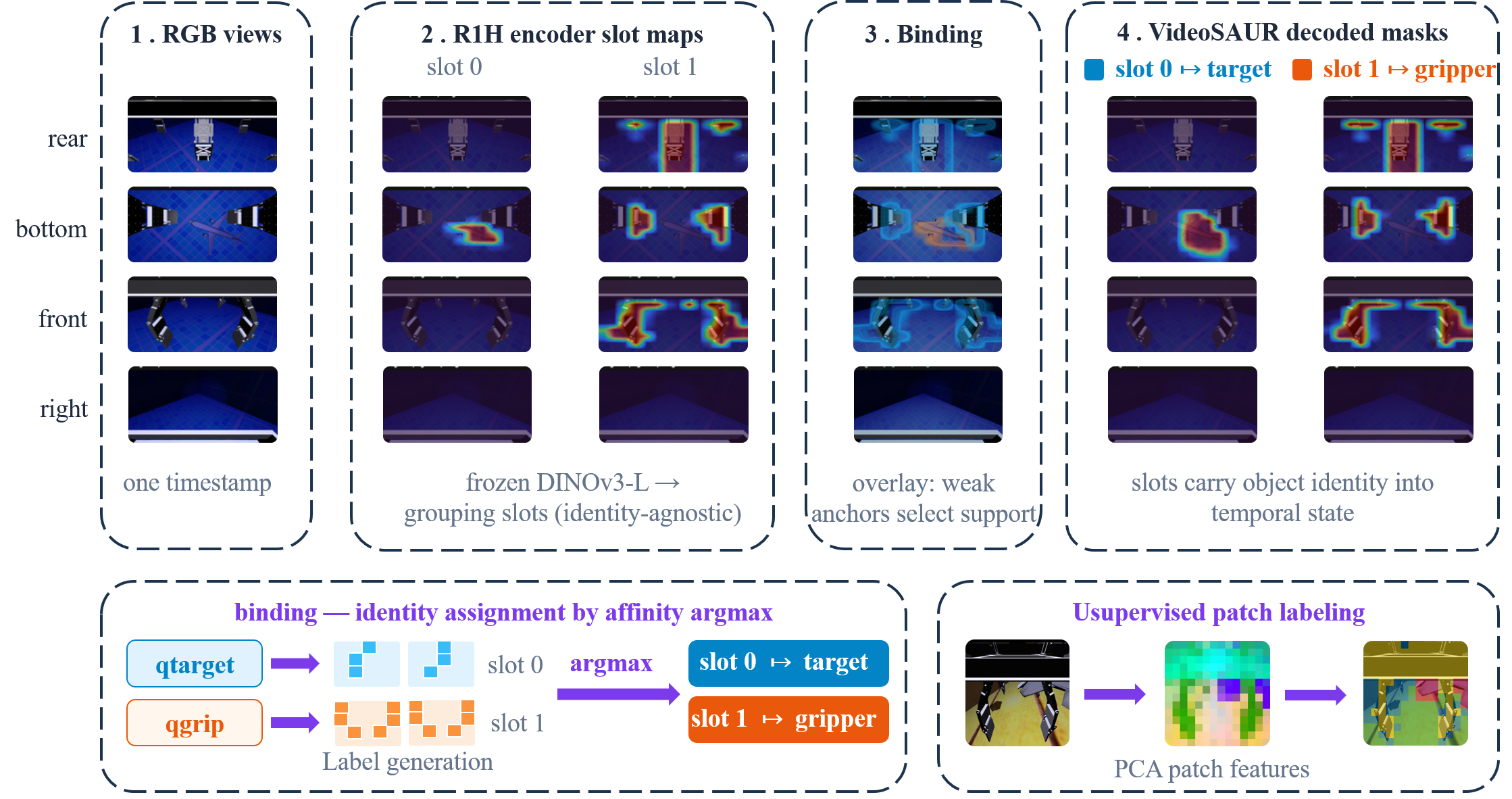}
\caption{Binding-guided patch-to-slot grounding in \method{}. Left to right: synchronized multi-view RGB frames from one real recording; grouping slot maps for the target (slot 0) and gripper (slot 1); the binding overlay, where weak anchors select patch support; the VideoSAUR-decoded slot masks that feed temporal state formation; and a PCA view of the patch features.}
\label{fig:binding_patch_to_slots}
\end{figure*}

For bound object $o$, a slot renderer produces a mask $\hat M_{t,o}^{v}$ and a weak mask proposal $M_{t,o}^{v}$ supplies an identity anchor. We use
\begin{equation}
\begin{split}
\mathcal{L}_{\mathrm{repr}} ={} & \mathcal{L}_{\mathrm{feat}}+\lambda_b\mathcal{L}_{\mathrm{bind}}+\lambda_m\mathcal{L}_{\mathrm{merge}} \\
& +\lambda_t\mathcal{L}_{\mathrm{timesim}}+\lambda_v\mathcal{L}_{\mathrm{vis}}+\lambda_s\mathcal{L}_{\mathrm{SIGReg}},
\end{split}
\end{equation}
where $\mathcal{L}_{\mathrm{feat}}$ reconstructs frozen DINOv3 patch features, free slots receive no object identity label, $\mathcal{L}_{\mathrm{timesim}}$ matches consecutive slot states to keep tracks temporally consistent, and $\mathcal{L}_{\mathrm{vis}}$ supervises the per-view visibility of the bound objects. The binding path is patch-to-slot rather than a detached DINO segmentation head: frozen DINOv3 patch evidence is scored against weak target/gripper anchors, and the selected support guides the corresponding VideoSAUR slot masks (Fig.~\ref{fig:binding_patch_to_slots}). Identity is assigned by affinity argmax against the query vectors $q^{\mathrm{target}}$ and $q^{\mathrm{grip}}$. A PCA--clustering view can still serve as a rapid proposal for the patch support. This makes the supervision auditable and inexpensive, but does not convert the method into fully unsupervised segmentation.

SIGReg is the Sketched Isotropic Gaussian Regularizer introduced for JEPA representations and used in LeWM to prevent collapse by matching the latent marginal to an isotropic Gaussian \cite{Balestriero_2025,Maes_2026_LeWM}. Let $Z\in\mathbb{R}^{N\times B\times d}$ collect token embeddings over a temporal window of length $N$, a batch of size $B$, and embedding dimension $d$. SIGReg projects $Z$ onto $M$ random unit directions $u^{(m)}\in\mathbb{S}^{d-1}$, applies the univariate Epps--Pulley normality statistic $T(\cdot)$ to each projection, and averages the result:
\begin{equation}
\mathcal{L}_{\mathrm{SIGReg}}(Z)=\frac{1}{M}\sum_{m=1}^{M}T\!\left(h^{(m)}\right),\qquad h^{(m)}=Zu^{(m)}.
\label{eq:sigreg_v35}
\end{equation}
The Cram\'er--Wold construction makes one-dimensional projections a tractable surrogate for the full distribution. SIGReg regularizes token geometry but does not assign target or gripper semantics; those come from weak binding and object-feature reconstruction.

\subsection{Held-out-view fusion}
For each bound object, cross-attention receives visible-view tokens and returns a fused token $\bar{s}_t^o$. During training, target view $v^*$ is excluded from keys and values. A view-conditioned decoder predicts its held-out feature evidence,
\begin{equation}
\mathcal{L}_{\mathrm{merge}}=\sum_o q_{t,o}^{v^*}\left\|D_{v^*}(\bar{s}_t^o)-X_{t,o}^{v^*}\right\|_2^2,
\label{eq:merge}
\end{equation}
where $q_{t,o}^{v^*}$ masks genuinely invisible objects. This test requires fused tokens to retain cross-view object evidence without requiring all camera appearances to collapse.

\subsection{Control-conditioned JEPA prediction}
The conditioning vector is the vehicle's normalized control signal,
\begin{equation}
c_t=[\tilde u_1,\ldots,\tilde u_{12},\tilde g_1,\ldots,\tilde g_6]_t,
\end{equation}
combining thruster command channels and gripper command channels. An MLP produces AdaLN-Zero parameters for the predictor \cite{Peebles_2023}. Given a history of four states, the model predicts twelve future states and minimizes token MSE plus cosine distance. The complete predictor has only $1.33$\,M parameters: because it predicts fused object tokens rather than pixels, the decision interface stays light even as the learned dynamics grow complex. In simulation the model is driven by commands rather than by measured motion, so the hydrodynamic mapping from control to realized vehicle motion, including thrust buildup, added-mass inertia, and cross-couplings, is learned inside the predictor. The control interface is modality-agnostic: thruster and gripper commands in simulation and recorded control and DVL-derived motion in the field drive the same autoregressive rollout. This makes the predictive state directly usable as a decision variable: candidate command sequences can be rolled out, scored, and replanned against the learned dynamics. Sec.~\ref{sec:mpc_interface} validates each ingredient of that loop separately. Extended architectural and training details are given in Sections~S2 and S3 of the supplementary material.

\section{Experimental Setup and Evaluation}\label{sec:setup}

Table~\ref{tab:config} lists the training configuration of the visual encoder and the control-conditioned predictor, shared by all experiments below.

\begin{table*}[t]
\centering
\caption{Training configuration of the visual encoder and the control-conditioned JEPA predictor.}
\label{tab:config}
\small
\begin{tabularx}{\textwidth}{@{}lX@{}}
\toprule
\multicolumn{2}{c}{\textbf{Visual encoder (frozen DINOv3-L backbone)}}\\
\midrule
Slots & 6 total: 3 binding (target UUV, gripper, and a binding-only slot for the fixed ROV frame), 3 free context \\
Slot dimension & $d_s=256$ (with a 256$\rightarrow$128$\rightarrow$256 adapter) \\
Fusion & held-out-view cross-attention, 8 heads (CrossViewMerge) \\
Reconstruction target & frozen DINOv3-L fp16 patch features \\
Loss weights & $\lambda_b{=}1.0$, $\lambda_m{=}1.0$, $\lambda_t{=}0.10$, $\lambda_v{=}0.20$, $\lambda_s{=}0.03$ \\
Training & lr $1\times10^{-4}$, $\approx$644 epochs, batch 16$\times$8 GPUs \\
\midrule
\multicolumn{2}{c}{\textbf{Control-conditioned JEPA predictor (transformer encoder)}}\\
\midrule
History / future frames & 4 / 12 (recorded as $1.0$ s / $3.0$ s of context) \\
Anchored masked slots & $1$ (one masked slot identity per window, recovered from the anchor frame) \\
Transformer & depth 6, width 128, 8 heads, GELU, norm-first, AdaLN-Zero \\
Optimizer & AdamW, lr $5\times10^{-4}$, weight decay $0.05$ \\
Epochs / batch / split & 30 / 256; 159 train, 18 validation \\
\midrule
\multicolumn{2}{c}{\textbf{Longer-context variants (stability probe)}}\\
\midrule
Context frames & 32 ($8.0$ s) and 16 ($4.0$ s), 25 epochs, initialized from the short-context predictor \\
\bottomrule
\end{tabularx}
\end{table*}

\subsection{Dataset taxonomy and split policy}
Data are collected in the ROS--Gazebo UUV Simulator environment \cite{Manhaes_2016_UUVSim}, which implements Fossen's equations of motion for the vehicle dynamics \cite{Fossen_2021}; each recording provides synchronized cameras, vehicle state, gripper signals, poses, and event annotations. The caveat of Sec.~\ref{sec:intro} about solving this gripper in simulation concerns closed-loop policy training; here the simulator instead supplies synchronized observations and privileged state, from which the model learns the command-conditioned mapping. Model inputs are the camera streams and command channels; privileged simulator state is used only to build training targets and evaluation references. The current camera configuration passes channel-wise attenuation values $(0.3,0.3,0.09)$ and RGB noise with standard deviation $0.02$, so the recordings embody a simplified attenuation/noise camera model rather than full radiative transfer. We package recordings by behavior rather than by storage folder: \emph{privileged automatic} data comprise wandering near a target, blind touching, and two naive privileged auto-grasp batches; \emph{manual teleoperation} data comprise wandering, gripper motion, interaction, grasping, and verification. The release used here contains 705 recordings across the two privileged auto-grasp batches, plus grasping (41), interaction (30) and gripper-motion (14) recordings; the remainder is unscripted collection. Each release manifest records source behavior, camera availability, calibration, annotation provenance, and splits. All results use recording-disjoint train/validation partitions, so no recording contributes frames to both sides of any reported comparison. Beyond the simulator, two field deployments of the same vehicle working a submerged real target supply synchronized camera streams and navigation telemetry; field evidence is reported in relative units in Sec.~\ref{sec:field}.

\subsection{Metrics and evidence ladder}
We evaluate five progressively stronger levels:
\begin{enumerate}
\item \textbf{Representation:} task-object mask Dice, held-out-view feature error, and slot stability.
\item \textbf{Prediction:} held-out token MSE and cosine distance under teacher forcing.
\item \textbf{Recursive rollout:} horizon-stratified comparison to persistence and controls shuffled within matched windows.
\item \textbf{Observable calibration:} position, relative-pose, or image-space error through a frozen readout with documented calibration limits.
\item \textbf{Decision and transfer:} executable-action ranking and conservative closed-loop simulation.
\end{enumerate}
An improvement at one level is not evidence for a later level. The full evaluation protocols, including the held-out-view protocol and the negative controls, are specified in Section~S4 of the supplementary material.

\section{Results}\label{sec:results}
The evidence below follows the evaluation ladder of Sec.~\ref{sec:setup}. 

\subsection{Representation evidence}
The learned representation preserves task-object semantics while reconstructing frozen visual features (Table~\ref{tab:representation}). Across the recording-disjoint split, binding-guided training maintains high task-group Dice for the target UUV and gripper while held-out-view fusion lowers feature reconstruction error.

\noindent\textbf{Binding guidance trades reconstruction for downstream utility.} Compared with an unguided self-supervised slot baseline (identical architecture and equal budget, but no binding anchors), binding guidance makes the slot decomposition semantically explicit---the target and gripper occupy stable, identifiable slots (group Dice $\approx0.90$/$0.88$) whereas the unguided baseline's slots carry essentially no task-object semantics (group Dice $\approx0.09$/$0.12$)---while feature reconstruction error rises only modestly, from $0.0157$ to $0.0186$ (Table~\ref{tab:representation}). The representation therefore pays a small reconstruction cost in exchange for task-relevant structure, confirming that reconstruction fidelity and downstream utility are distinct objectives.

\noindent\textbf{Held-out-view fusion, measured against matched arms.} Block (b) of Table~\ref{tab:representation} isolates the fusion design on a recording-disjoint split at equal budget, where the arms differ only in the flags named in the row. The two blocks are not comparable in absolute level, and all between-arm claims are drawn from (b). Replacing weighted-sum fusion with held-out-view cross-attention improves \emph{all five} axes: feature MSE falls from $0.0231$ to $0.0200$ ($-13.4\%$), target decoder Dice rises from $0.713$ to $0.820$ ($+10.7$\,pp), gripper decoder Dice from $0.814$ to $0.891$ ($+7.7$\,pp), and the target and gripper group Dice by $+2.7$ and $+3.8$\,pp. Two further contrasts sharpen what is doing the work. Against averaging the visible views with the target held out, cross-attention is better on four of five axes (target group $0.821\to0.826$, target decoder $0.811\to0.820$, gripper group $0.836\to0.843$, feature MSE $0.0201\to0.0200$). And against the \emph{same} cross-attention allowed to attend to the target view, holding that view out is again better on four of five axes (target group $0.796\to0.826$, gripper group $0.824\to0.843$, target decoder $0.805\to0.820$, feature MSE $0.0205\to0.0200$), consistent with the constraint acting as a regularizer. One axis runs the other way: gripper decoder Dice is $1.1$--$1.2$\,pp lower under held-out fusion ($0.891$ against $0.902$/$0.903$). The rear-view-only arm is not a competing fusion method: it is trained with the target view included and cannot answer Equation~\eqref{eq:merge}; its comparable average Dice contributes nothing to the held-out claim. The block~(b) margins come from a single training run per arm and are read as effect sizes rather than as variance estimates.

\noindent\textbf{The fused token carries evidence for a view it never saw.} The ladder above measures average quality; the direct test is whether the fused token can reconstruct a withheld view at all. On a purpose-built held-out evaluation of the same design (each of the four cameras held out in turn; the zero-information floor averages the three visible views' local outputs), held-out target Dice rises from $0.2206$ to $0.7604$ for the UUV and from $0.2034$ to $0.8695$ for the gripper, held-out feature MSE falls from $2.3104$ to $1.2183$ and from $2.3552$ to $1.0148$, and spurious mask mass on frames where the target is absent in that view collapses from $0.1339$ to $0.0063$ and from $0.1257$ to $0.0007$. The shared token does not collapse: its effective rank \cite{Garrido_2023_RankMe} is $14.69$ for the UUV and $8.71$ for the gripper, out of $128$ dimensions. Two limits belong with this result. The evaluation uses $512$ fixed random windows rather than a recording-disjoint split, so it establishes the mechanism and its magnitude, not a generalization estimate; and four of its five acceptance criteria, fixed before the evaluation, pass, with the front camera (which meets the target at a grazing angle) landing at $0.456$ UUV Dice where right, bottom and rear sit between $0.755$ and $0.800$. That spread is itself the argument for fusion over any single chosen view: which view will be the inadequate one is not knowable in advance.

\noindent\textbf{SIGReg improves the task-object position head.} Separately from the DINOv3 checkpoint in Table~\ref{tab:representation}, a representation sweep isolated the contribution of SIGReg by training the geometry renderer with and without the SIGReg term while keeping everything else fixed. Adding SIGReg at weight $0.03$ reduced the estimated 3D target-center error from $0.625$ to $0.496$ (normalized center error), increased Dice from $0.887$ to $0.895$, and improved both IoU and feature reconstruction. The position head therefore benefits from SIGReg independently of the fusion change: a small amount of latent-geometry regularization sharpens the geometry head and compounds with, rather than substitutes for, the object-centric supervision.

\begin{table*}[t]
\centering
\caption{Representation checkpoints. Higher Dice and lower feature MSE are better. \textbf{(a)} Deployed full-data checkpoints; the unguided row is the equal-budget baseline without binding anchors. \textbf{(b)} Equal-budget fusion ladder.}
\label{tab:representation}
\small
\setlength{\tabcolsep}{3pt}\begin{tabular}{@{}>{\raggedright\arraybackslash}p{3.3cm}ccccc@{}}
\toprule
\multicolumn{6}{c}{\textbf{(a) Deployed checkpoints}}\\
\midrule
Model & Feat. MSE & Target group & Target dec. & Grip group & Grip dec.\\
\midrule
Binding + SIGReg & 0.0186 & 0.897 & 0.915 & 0.881 & 0.933\\
Held-out fusion & 0.0183 & 0.895 & 0.903 & 0.877 & 0.921\\
Unguided self-supervised (no binding) & 0.0157 & 0.085 & 0.018 & 0.116 & 0.118\\
\midrule
\multicolumn{6}{c}{\textbf{(b) Equal-budget fusion ladder}}\\
\midrule
Binding, weighted-sum fusion (no cross-view) & 0.0231 & 0.799 & 0.713 & 0.805 & 0.814\\
Binding, rear view only & 0.0201 & 0.828 & 0.811 & 0.845 & 0.902\\
Binding, mean of visible views, target held out & 0.0201 & 0.821 & 0.811 & 0.836 & 0.902\\
Binding, cross-attention, target view exposed & 0.0205 & 0.796 & 0.805 & 0.824 & 0.903\\
Binding, cross-attention, target held out (\gocvf{}) & $\mathbf{0.0200}$ & $\mathbf{0.826}$ & $\mathbf{0.820}$ & $0.843$ & $0.891$\\
\bottomrule
\end{tabular}
\end{table*}

\subsection{Conditional rollout evidence}
\begin{figure*}[t!]
\centering
\includegraphics[width=0.76\textwidth]{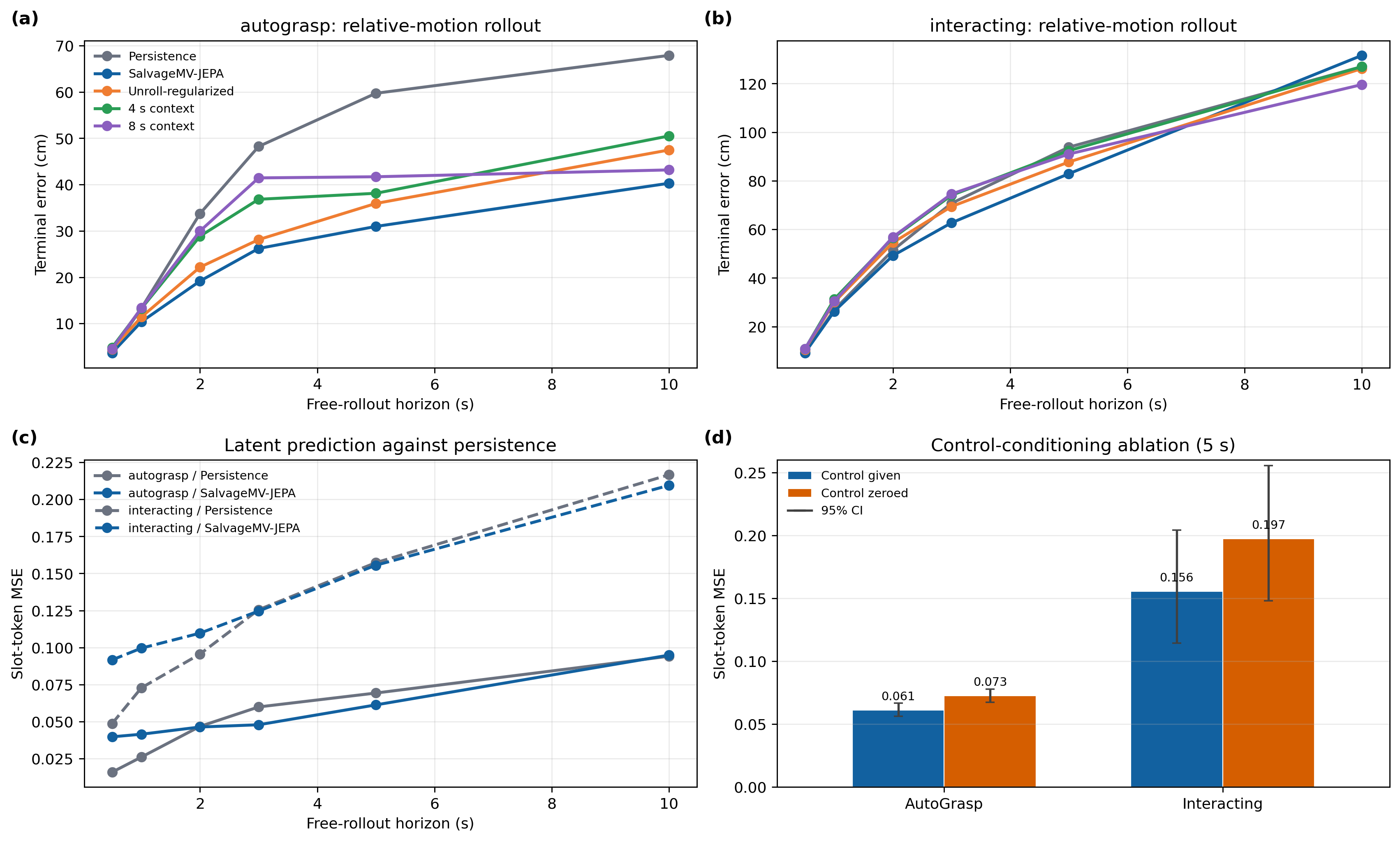}
\caption{Conditional rollout diagnostic on the 229-recording interaction set. (a,b) Terminal relative-motion error versus free-rollout horizon for persistence, \method{}, the unroll-regularized variant, and the long-context variants ($16/32$ frames); (c) latent MSE against persistence; (d) 5\,s control-conditioning ablation, recorded versus zeroed control, with bootstrap 95\% CIs.}
\label{fig:rollout_summary}
\end{figure*}

On a 229-recording interaction-focused diagnostic set, the conditional predictor reduced a frozen-readout terminal relative-motion error from 39.1 cm for persistence (recording-level bootstrap 95\% CI 36.7--41.5) to 25.7 cm (23.6--27.9). At five seconds, the same diagnostic reported 59.7 to 31.0 cm on the AutoGrasp recordings and 93.8 to 82.9 cm on the Interacting recordings. Zeroing the conditioning signal increased latent MSE, confirming that the learned tokens encode control-correlated change and outperform persistence in dynamic windows. Shuffling or zeroing the control input at a fixed state is an established intervention for verifying that a predictor conditions on actions rather than on visual habit \cite{Hang_2026_Habit}. The same diagnostic shows where persistence stays competitive (Fig.~\ref{fig:rollout_summary}c): at short horizons the model's latent MSE is comparable to persistence, the two curves nearly coinciding on AutoGrasp at 10\,s. A persistence copy is a strong latent-space baseline because most tokens change slowly; the model's advantage lies in the readout-relevant terminal error and in the control-conditioning margin, not in raw latent distance. Repeating the last observation as a latent-space baseline is standard practice rather than a strawman \cite{Zhang_2026_ThinkJEPA}, and recent work normalizes rollout error by a temporal-persistence baseline to discount temporal smoothness \cite{Wen_2026_JEPAx}. Longer-context variants (4\,s and 8\,s) and an unroll-regularized variant were probed on the same diagnostic; neither improves on the deployed 1\,s-context model overall (terminal error $32.2$\,cm for both longer-context variants, $28.6$\,cm for the unroll variant, versus $25.7$\,cm; Fig.~\ref{fig:rollout_summary}), the single exception being the 10\,s horizon on the Interacting recordings, where the 8\,s variant is best. The compact configuration is retained.

\noindent\textbf{Synthetic-perturbation surprise.} To test whether the predictor holds expectations about observation continuity rather than extrapolating means, we inject four perturbations into the history window and leave the target unchanged ($12{,}000$ windows over $250$ held-out records). Overwriting the history with the training-set mean token raises future error by $536\%$ ($95\%$ CI of the absolute increase $0.20$--$0.27$, detected in $82\%$ of windows); replacing it with a distant window of the same recording raises it by $260\%$ ($0.10$--$0.12$, $75\%$). Reversing the three pre-anchor history frames or permuting them moves the prediction by $1.6\%$ and $0.6\%$. The predictor is therefore sharply sensitive to violations of observation continuity, and nearly indifferent to frame ordering inside a one-second history.

\begin{figure}[!b]
\centering
\includegraphics[width=\columnwidth]{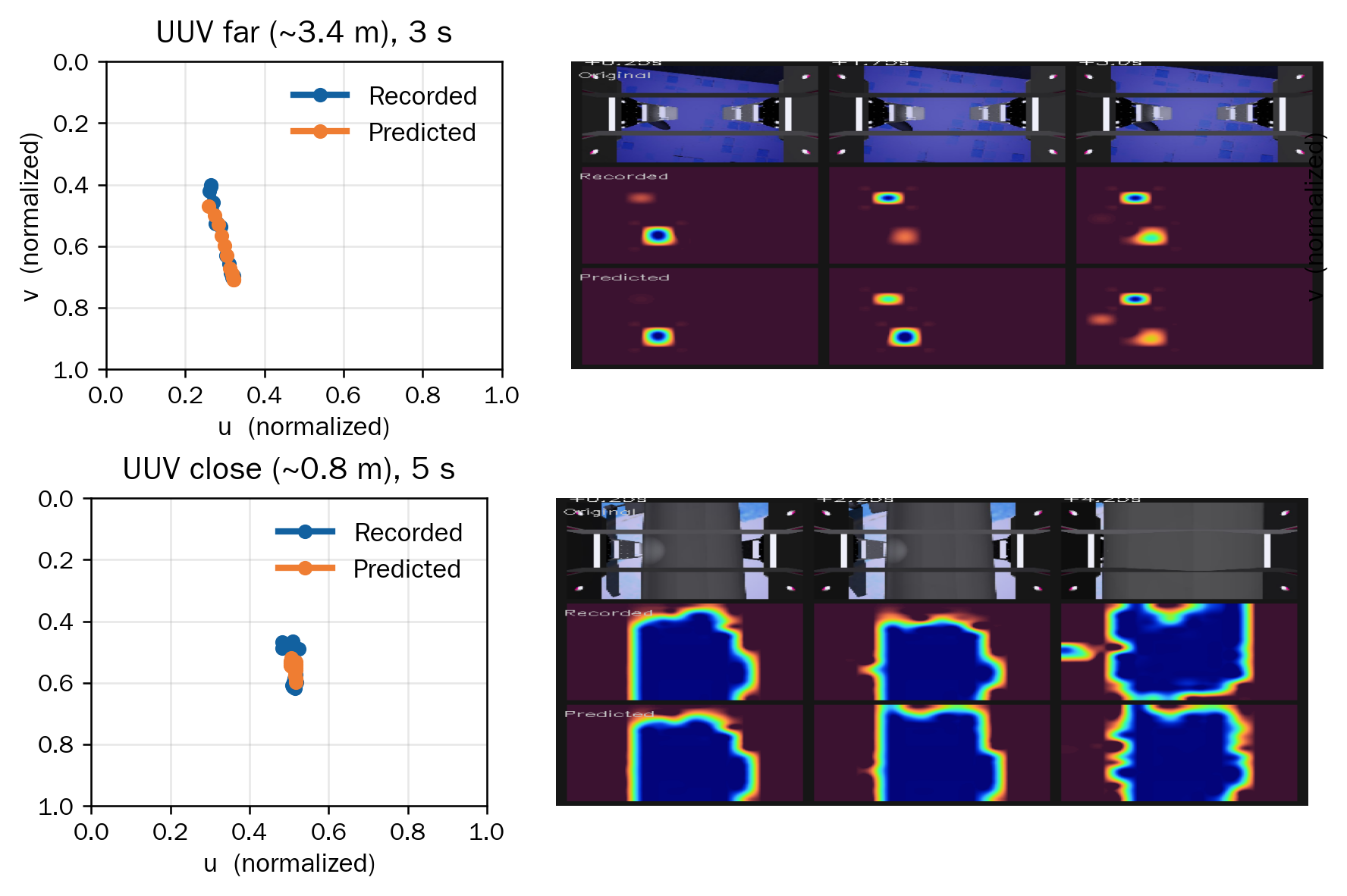}
\caption{UUV (slot 0) mask motion under the world model at two ranges: far ($\approx$$3.4$\,m, 3\,s rollout, top) and close ($\approx$$0.8$\,m, recursive 5\,s, bottom). Each panel pairs recorded (blue) and predicted (orange) mask-centroid tracks with the bottom-camera frame.}
\label{fig:uuv_motion}
\end{figure}

Using the frozen geometry UUV renderer, we also decode the predicted future UUV token into a slot mask and compare its centroid track to the recorded one (Fig.~\ref{fig:uuv_motion}). The world model reproduces the recorded UUV motion both when the target is far from the ROV ($\approx$$3.4$\,m, over a 3\,s window the recorded mask center moves by about $1.1$ normalized units and the predicted center tracks it with a mean error of about $0.09$) and when the target is close ($\approx$$0.8$\,m), where a recursive 5\,s rollout keeps the mean centroid-tracking error under $0.05$ normalized units with no divergence at the endpoint.  The multi-view design is what makes this tracking robust in near-field salvage: as a task object moves out of one camera's field of view and into another, a single camera would lose the object exactly when its view is lost, while held-out-view cross-attention aggregates the evidence still visible in the other cameras. Consistent with this mechanism, the shared target token is decoded to the same 3D location across views (its mask-centroid rays stay close, including for the \emph{predicted} token; see Fig.~\ref{fig:uuv_motion} and the cross-view results below), and the observed trajectory is preserved through an imagined rollout.

\subsection{Cross-view consistency}
A shared task token must fuse evidence across views: the masks it decodes in different cameras should point at the same 3D location. We measure the mean pairwise angular spread of the four view-centroid rays for the UUV token on 15 interacting recordings, binned by prediction horizon, and compare it between the world model's predicted token and the observed token. The two agree closely at every horizon ($16$--$20^\circ$ predicted versus $17$--$22^\circ$ observed), and the mean predicted spread is no larger than the observed spread within each bin. Prediction thus preserves the multi-view fusion structure: the fused token stays view-consistent through an imagined rollout.

\subsection{Causal consistency and behavior-agent training}
Beyond representation, we test whether conditioning is real: the predicted slot motion should track the true motion only when the recorded control is supplied. On the same interacting set, the direction cosine between the predicted and true UUV token displacement rises to about $0.42$ with control at 15\,s versus $0.23$ without control (recording-weighted means over the phase-split rows of Table~S4 of the supplementary material; the simple average of the two 15\,s rows differs because weighting follows per-phase counts), so the model uses the conditioning signal. Splitting by the gripper command, the grasp/contact phase shows a positive control gain at every horizon, growing from $+0.07$ at 1\,s to about $+0.24$ at 15\,s; the approach phase is positive only at the longer horizons (slightly negative at 1\,s, $+0.06$ at 5\,s, up to about $+0.16$ at 10\,s). Table~S4 of the supplementary material reports the recording-level means.

We further probe whether the world model can \emph{train} a behavior agent. Using the conditional dynamics and a frozen potential critic, we train an action-generating agent with behavior-cloning regularization under imagined rollouts, separately in the search and grasp/contact phases. On held-out recordings the learned agent's critic-predicted task progress exceeds the behavior-cloning baseline by $+0.383$ (search) and $+0.209$ (grasp/contact). The imagination interface guides agent improvement in both phases. Progress is measured by the same frozen critic that shapes the training signal, so this is a statement about the consistency of the imagined-rollout loop rather than an independent task-success measure.

\subsection{Interface validation for model-predictive control}
\label{sec:mpc_interface}
The matched-endpoint ranking diagnostic validates the planning interface: candidate trajectories share a common terminal pose and differ only in control timing, and the world model's predicted latent endpoints rank them well above shuffled-control rollouts. Concretely, on 90 matched-endpoint candidate groups under recording-level out-of-fold evaluation, ranking by the world model's predicted endpoint selects the true best candidate first in $53.3\%$ of groups (bootstrap 95\% CI $43.1$--$63.3\%$) versus $23.3\%$ ($15.8$--$33.1\%$) for shuffled-control rollouts and $25\%$ for random choice, with mean regret (selected-candidate distance penalty relative to the group's best) reduced from $0.0091$ to $0.0058$; each group contains four trajectories, so chance is $25\%$, and the model's interval excludes both chance and the shuffled-control rate. Recent work audits whether latent distances rank candidate plans consistently with executed outcomes \cite{Wang_2026_DMA}; the diagnostic above is a positive instance. Together with the causal-consistency gains of Table~S4 of the supplementary material (positive at every horizon in the grasp/contact phase), all three ingredients of the model-predictive loop are in place: a control-conditioned predictor, a consistent latent geometry, and a ranking signal that separates candidate actions. Closed-loop operation composes these verified ingredients with a conventional automatic controller, as described in Sec.~\ref{sec:field}.

\subsection{Downstream transfer against a reconstruction-free world model}
\label{sec:lewm_compare}
To quantify how much task-relevant information each representation retains, we train a reconstruction-free latent world model (LeWM \cite{Maes_2026_LeWM}: ViT encoder, latent next-step prediction, SIGReg prior on the embedding; no reconstruction, no slots, no object supervision) from scratch on the same multi-view recordings and control signals as \method{}, and compare frozen representations under an identical probe architecture, split, and optimization predicting gripper joint angles, target relative position, and target yaw. Table~\ref{tab:lewm_compare} reports the result: \method{} roughly halves the joint-angle and yaw error of the reconstruction-free embedding and reduces relative-position error by a quarter, transferring substantially more task information on every head.

A purely latent objective with SIGReg as its only regularizer discards a large share of the visual information manipulation requires, driving the embedding toward compressed dynamics statistics and away from pose- and joint-level evidence. The object-centric supervision in \method{}---reconstruction-guided slots with weak binding and a small SIGReg weight ($0.03$)---is what preserves that evidence. SIGReg alone does not recover it, and at reconstruction-free training scales it actively hurts downstream heads; combined with reconstruction and binding, the same regularizer improves them.

\begin{table}[!ht]
\centering
\caption{Downstream probe transfer on the grasping scene (recording-disjoint split; identical probe architecture and training). Lower is better. LeWM is trained from scratch on the same recordings and control signals as \method{}.}
\label{tab:lewm_compare}
\footnotesize
\setlength{\tabcolsep}{2pt}\begin{tabular}{@{}>{\raggedright\arraybackslash}p{2.3cm}ccc@{}}
\toprule
Representation & Joint MAE ($^\circ$) & Rel.\ pos.\ MAE (m) & Yaw MAE ($^\circ$)\\
\midrule
\method{} (binding + slots + SIGReg) & $\mathbf{4.90}$ & $\mathbf{0.60}$ & $\mathbf{6.50}$\\
LeWM (latent, SIGReg only) & 9.81 & 0.79 & 13.82\\
\bottomrule
\end{tabular}
\end{table}

\section{Field Validation on Real Underwater Video}
\label{sec:field}

\begin{figure*}[t!]
\centering
\includegraphics[width=0.70\textwidth]{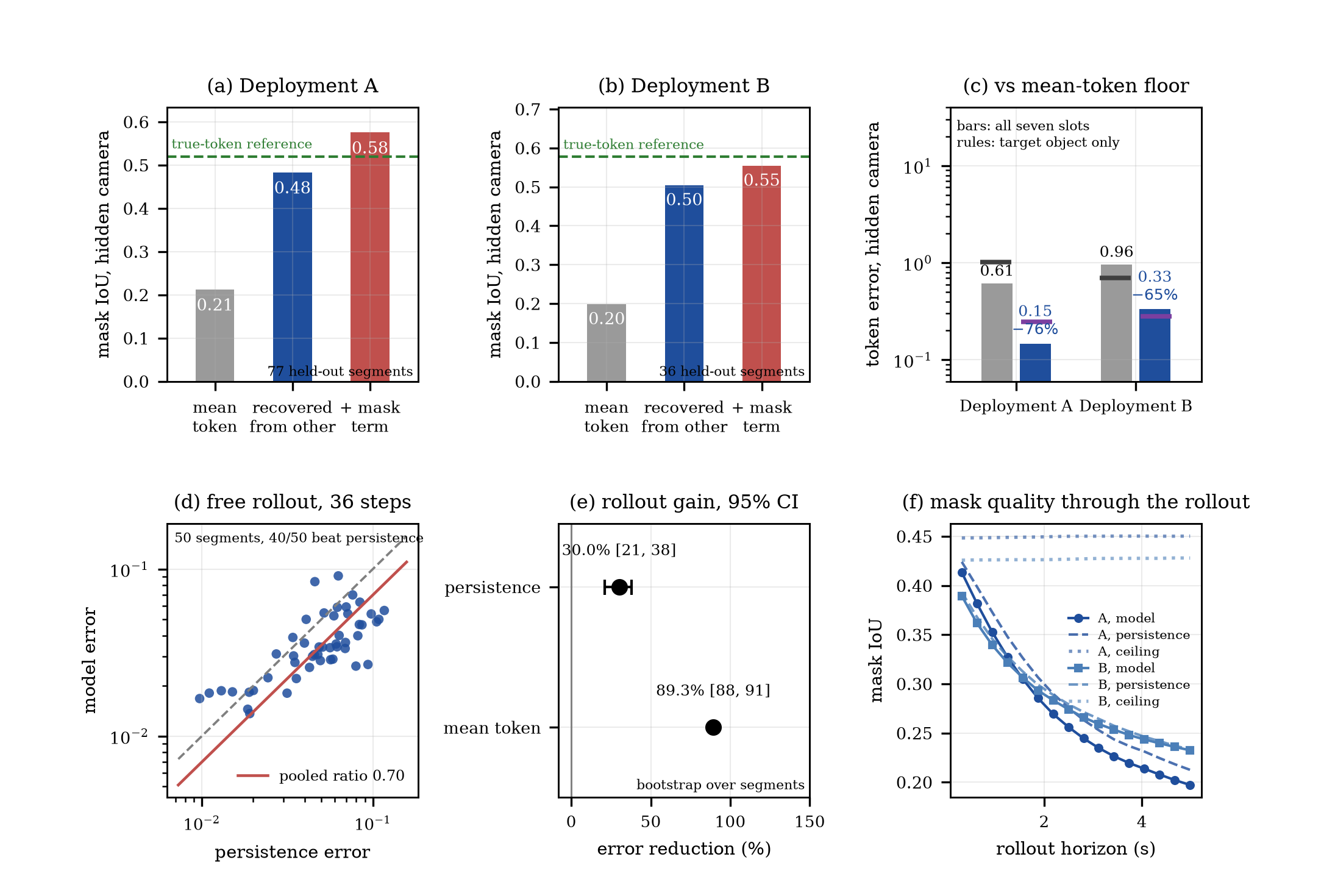}
\caption{Field validation. (a,b) Mask quality on the withheld camera for the mean-token, recovery and mask-term arms; dashed: that camera's true tokens. (c) Token error of that recovery against the mean-token floor, over the seven slots (bars) and the target (rules). (d) Per-segment token error of a 36-step free rollout against persistence. (e) Error reduction over each reference. (f) Mask quality through the rollout.}
\label{fig:field}
\end{figure*}

All results above are trained and measured on simulated salvage scenes. The question here: does the same predictive state, with the same recipe, hold up on real underwater video? Everything here is trained from the raw field recordings alone; the simulator does not participate. Open-loop replay of recorded actions is a recognized evaluation protocol when closed-loop deployment is unavailable \cite{Aljalbout_2026_RealityGap}.
Both recordings were collected in a test pool measuring 60\,m by 20\,m with a water depth of 10\,m, under two operation modes (Fig.~\ref{fig:conditions}): unscripted navigation 3--5\,m above a steel frame installed on the pool floor, and active interaction with an aluminum pipe on the floor, bringing the gripper into contact with the object. The two deployments are reported separately and never pooled: Deployment A is the navigation recording, which keeps the steel frame in view; Deployment B is the contact recording. The platform is the heavy-duty salvage ROV of Fig.~\ref{fig:rov_platform}, with paired linkage-driven underactuated grippers, a synchronized camera pair, navigation telemetry, and no contact sensing. Both are recorded by two co-mounted cameras for which no intrinsics or extrinsics were measured, and the target carries no metric ground truth. A video walkthrough of the experiments accompanies this paper as supplementary multimedia.

\begin{figure*}[t!]
\centering
\begin{tabular}{@{}cccc@{}}
\includegraphics[height=2.8cm]{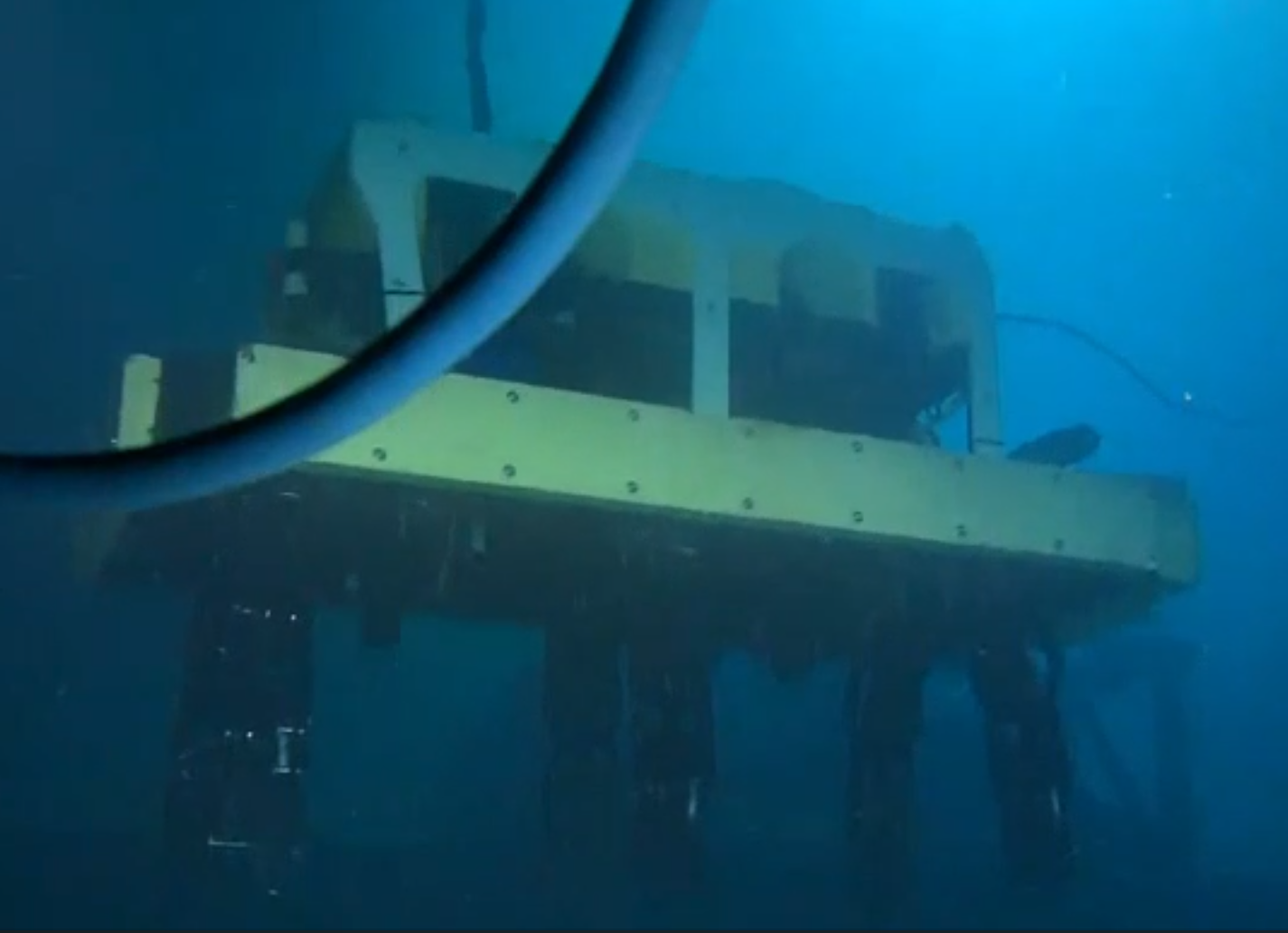} & \includegraphics[height=2.8cm]{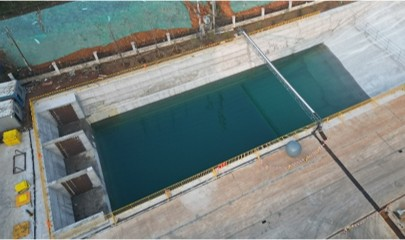} & \includegraphics[height=2.8cm]{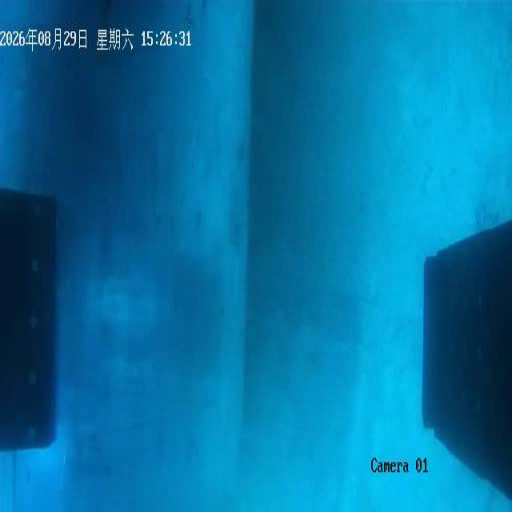} & \includegraphics[height=2.8cm]{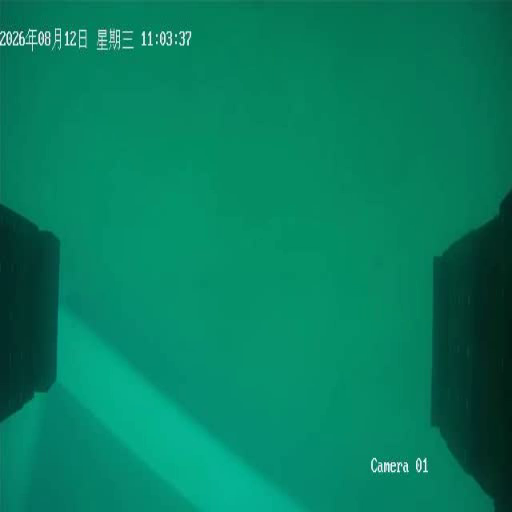}\\[1pt]
{\small (a)} & {\small (b)} & {\small (c)} & {\small (d)}
\end{tabular}
\setcounter{figure}{8}
\caption{Field experimental conditions. (a) The heavy-duty underwater salvage ROV used in this study. (b) The test pool: 60\,m by 20\,m, 10\,m deep. (c) The aluminum pipe on the pool floor. (d) The steel frame on the pool floor.}
\label{fig:conditions}
\setcounter{figure}{6}
\end{figure*}

Three reference points appear throughout: predicting the mean token of the training sessions, copying the previous token (persistence), and decoding the true tokens of the view being predicted. Token error is the per-dimension mean squared error against these references, and mask quality is reported as a floor/model/ceiling intersection-over-union triple. 

\begin{figure}[t!]
\centering
\includegraphics[width=\columnwidth]{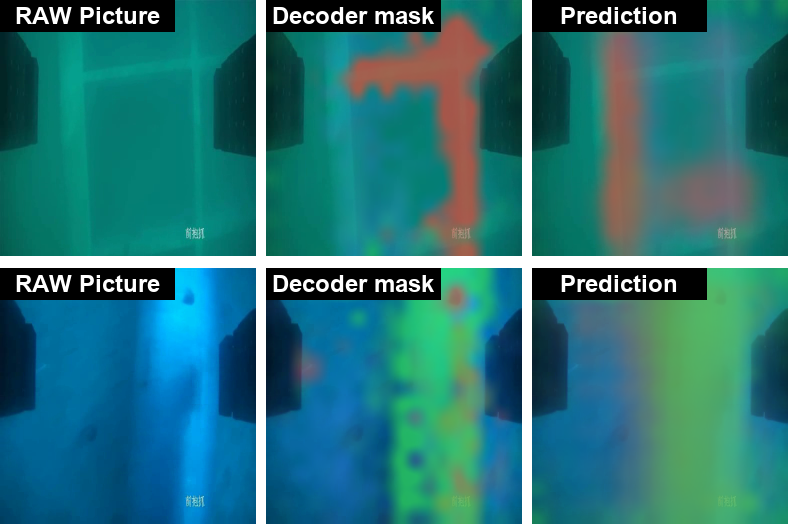}
\caption{Hidden-camera reconstruction on the two deployments (one row each):
the field frame from the withheld camera, the mask decoded from its own tokens, and the mask recovered from the other camera.}
\label{fig:field_qual}
\end{figure}

\begin{figure}[t!]
\centering
\includegraphics[width=\columnwidth]{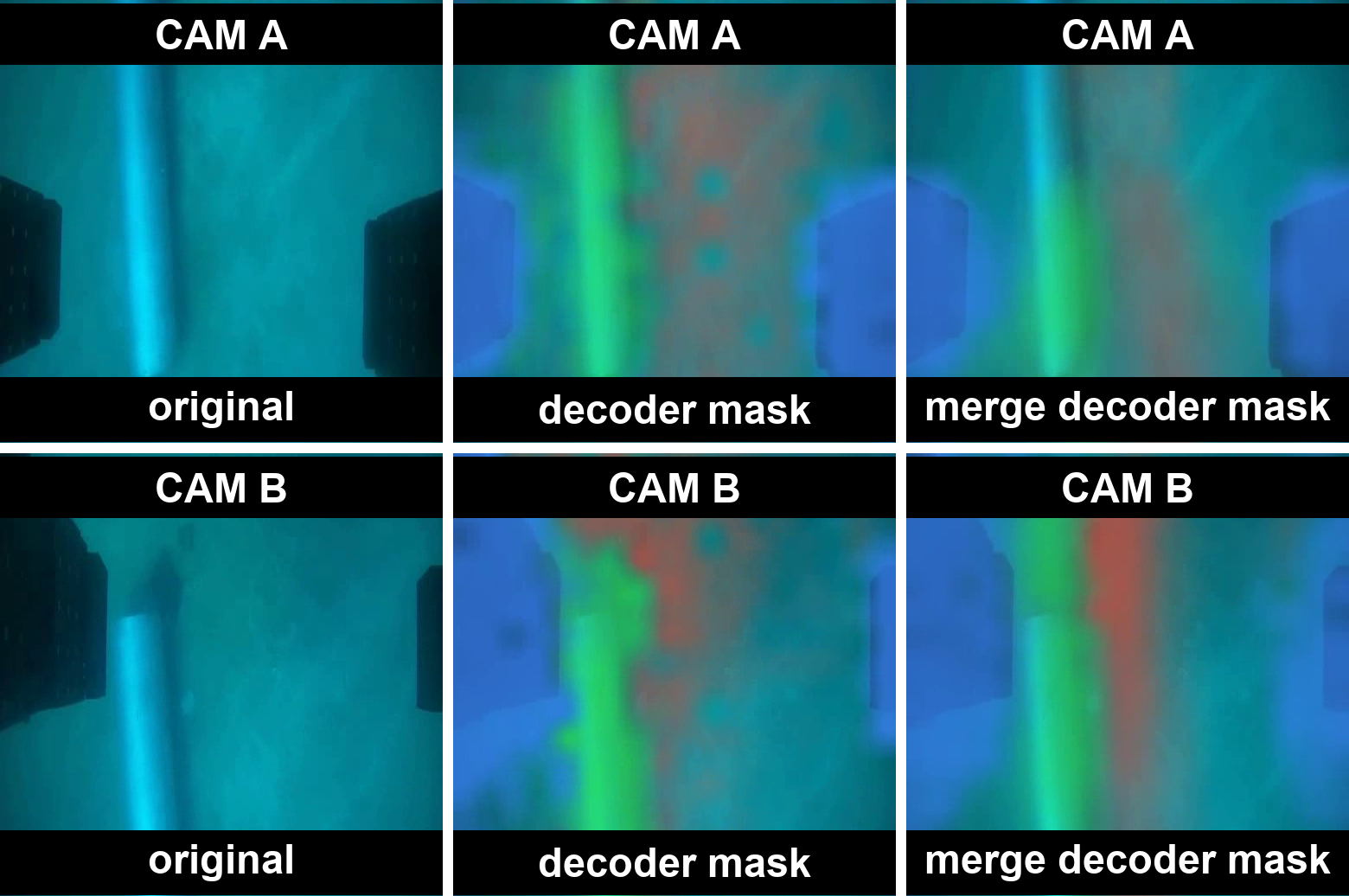}
\caption{Cross-view decoding on a field recording (CAM~A top row, CAM~B bottom
row): the original frame, the mask from that camera's own tokens, and the merge decoder mask from cross-view fusion.}
\label{fig:fused}
\end{figure}

\noindent\textbf{Evidence for a hidden camera survives in the shared tokens.}
Predicting one camera's tokens from the other camera alone lowers token error
by $76\%$ and $65\%$ on the two deployments and recovers the hidden camera's
mask at $88\%$ and $81\%$ of the achievable range
(Table~\ref{tab:field_summary}; Fig.~\ref{fig:field}a--c and
Fig.~\ref{fig:field_qual}). The recovered tokens also keep their spread, close
to the true spread against about half of it for the mean-token prediction, so the gain is not the trivial mean-token shrinkage effect. A linear readout of the same history fails: it is $5.7\times$ and $3.5\times$ worse
than persistence on the two deployments, and once targets are expressed as frame-to-frame change it is
indistinguishable from a shuffled control. The evidence is present but not linearly recoverable; a nonlinear head is therefore required. The two
cameras are rigidly co-mounted and largely co-visible, so the field study tests
hidden-view recovery (the capability fusion must deliver) rather than a
dedicated fusion benchmark; the identity-preserving form of the fusion claim
is established on the simulated camera rig of Sec.~\ref{sec:results}. Fig.~\ref{fig:fused} shows both cameras of one recording decoded from their own tokens and from the merged cross-view tokens.

\begin{table}[!ht]
\centering
\caption{Field-validation summary. Deployment A: navigation above a submerged steel frame; Deployment B: contact with a submerged aluminum pipe.}
\label{tab:field_summary}
\small
\setlength{\tabcolsep}{3pt}
\begin{tabular}{@{}lcc@{}}
\toprule
 & Deployment A & Deployment B\\
 & (77 segments) & (36 segments)\\
\midrule
Hidden-view token error & $0.615 \to 0.147$ & $0.962 \to 0.332$\\
Hidden-view mask IoU & $0.21$ / $0.48$ / $0.52$ & $0.20$ / $0.50$ / $0.58$\\
Recovered share of range & $88\%$ & $81\%$\\
$5$\,s prediction vs.\ persistence & $-23\%$ & $-31\%$\\
\bottomrule
\end{tabular}
\end{table}

\noindent\textbf{Prediction against recorded motion.} A state-only closed-form ridge regression on the $16$-step
token history reduces $5$\,s-ahead token error by $23\%$ and $31\%$ on the two
deployments relative to persistence (Table~\ref{tab:field_summary}); re-running
the reference implementation reproduces the recorded values to within
$4\times10^{-6}$. Letting the field-trained \method{} predictor, conditioned on the
recorded control signals, feed itself for 36 steps is the harder test, and it holds: error falls $30\%$ below persistence
($95\%$ CI $21$--$38\%$) and $89\%$ below the mean-token floor
(Fig.~\ref{fig:field}d--e).

\noindent\textbf{Mask quality follows its own axis.} The same
rollout that cuts token error by $30\%$ does not improve mask quality: $0.159$
against $0.184$ for persistence (CI $-0.053$ to $+0.002$,
Fig.~\ref{fig:field}f). Squared error on tokens is minimized by a shrunken
conditional mean, and the frozen token-to-mask decoder turns that shrinkage
into a diffuse mask that a fixed threshold penalizes. Training the recovery
with a mask term of weight $0.3$ lifts held-out mask quality by $0.092$ and
$0.050$ on the two deployments, at a token-error cost of $0.016$ and $0.032$.
A variant trained on masks alone draws masks as good as any other ($0.58$)
while its token error sits at twice the mean-token floor. Mask quality and
predictive fidelity are separate axes; we treat predictive fidelity as the
acceptance criterion and report mask quality as a floor/model/ceiling triple.

\noindent\textbf{Scope of the field evidence.} Field quantities are reported in relative units, and actuated behavior is exercised only in simulation. The withheld view is recovered close to its achievable limit, and the control-conditioned predictor stays ahead of persistence at planning horizons and over self-fed rollouts; per-camera breakdowns and the mask-term arms appear in Section~S6 of the supplementary material.

\method{} is a predictive state model. Binding keeps
the target and the gripper in stable slots, the fusion head gathers what the
cameras see, the remaining slots hold unmodeled context, and the predictor hands
state, predicted consequences and candidate scores to a decision module. In
simulation, coupling the interface to the platform's conventional controller
completes the approach, alignment, contact and hold sequence, and the separation
between predictive interface and control back end keeps the model portable
across platforms. We report this coupling as an interface demonstration, not a powered success-rate study; a closed-loop field trial is the natural next step. At deployment scale the predictor remains small: $1.33$\,M parameters, $6.5$\,M FLOPs per window, and $1.97$\,ms per forward pass on a single A800 GPU ($0.6\%$ of the $0.31$\,s state step), with a $76$\,MB peak footprint; the latency is a datacenter-GPU reference, and an embedded port of the same predictor is left to future work. Up to Sec.~\ref{sec:results} all evidence is produced in ROS--Gazebo with the simplified attenuation/noise camera model of Sec.~\ref{sec:setup}; the field study extends the evaluation to real underwater video.

\section{Conclusion}

We introduced \method{}, which instantiates the object-centric perception domain (\ocpd{}) of Sec.~\ref{sec:intro} through guided object-centric cross-view fusion (\gocvf{}) and predicts its evolution with a control-conditioned JEPA world model, for physics-based near-field operation of a heavy-duty underwater salvage ROV with an underactuated, self-adaptive gripper. The model combines low-cost weak binding, held-out-view fusion, free context slots, and control-conditioned latent prediction, internalizing the command-to-motion hydrodynamics without contact sensing. The representation preserves task-object masks through dynamic approach and contact windows (group Dice $\approx0.9$ versus $\approx0.1$ for unguided slots) and predicts the evolution of the object--gripper interaction, cutting persistence's terminal motion error by a third on the interaction diagnostic and by half at 5\,s on the AutoGrasp recordings. Binding guidance improves downstream target-pose estimation even at a higher reconstruction cost, while a small SIGReg weight independently sharpens the geometry head. Against a reconstruction-free latent world model trained on identical data, \method{} transfers substantially more task-relevant information to identical probes, showing that object-centric reconstruction guidance preserves information that latent regularization alone discards. The predictive interface supports MPC candidate rollouts, candidate ranking, and imagined-rollout behavior-agent training, and in simulation it is consumed by a conventional automatic controller. On real underwater video the same architecture reconstructs a withheld camera to 81--88\% of its achievable range and beats persistence by 23--31\% at a five-second horizon. More broadly, the method provides a compact recipe for constructing MPC-compatible world models for low-cost embodied agents with multiple sensing modalities, and the perception-domain view is what keeps that reference affordable: alignment is spent on task objects, and context is carried unaligned.

\section*{Acknowledgment}
This work was supported by the National Key Research and Development Program of China under Grant 2023YFC2809701. During the preparation of this work the authors used an AI-assisted writing tool (Hermes) to improve language and consistency; the authors reviewed and edited the content as needed and take full responsibility for the publication. The data and code that support the findings of this study are available from the corresponding authors upon reasonable request. The authors declare no conflicts of interest.

\bibliographystyle{IEEEtran}
\bibliography{refs}

@article{Simetti_2021,
  author    = {Enrico Simetti},
  title     = {Autonomous Underwater Intervention},
  journal   = {Current Robotics Reports},
  volume    = {2},
  pages     = {119--130},
  year      = {2021},
  doi       = {10.1007/s43154-021-00046-3}
}

@article{Sun_2024,
  author    = {Linxiang Sun and Yu Wang and Xiaolong Hui and Xibo Ma and Xuejian Bai and Min Tan},
  title     = {Underwater Robots and Key Technologies for Operation Control},
  journal   = {Cyborg and Bionic Systems},
  volume    = {5},
  pages     = {0089},
  year      = {2024},
  doi       = {10.34133/cbsystems.0089}
}

@article{Morgan_2022,
  author    = {Edward Morgan and Ignacio Carlucho and William Ard and Corina Barbalata},
  title     = {Autonomous Underwater Manipulation: Current Trends in Dynamics, Control, Planning, Perception, and Future Directions},
  journal   = {Current Robotics Reports},
  volume    = {3},
  number    = {4},
  pages     = {187--198},
  year      = {2022},
  doi       = {10.1007/s43154-022-00089-2}
}

@book{Fossen_2021,
  author    = {Thor I. Fossen},
  title     = {Handbook of Marine Craft Hydrodynamics and Motion Control},
  edition   = {2nd},
  publisher = {Wiley},
  year      = {2021},
  isbn      = {978-1119575054}
}

@inproceedings{Locatello_2020,
  author    = {Francesco Locatello and Dirk Weissenborn and Thomas Unterthiner and others},
  title     = {Object-Centric Learning with Slot Attention},
  booktitle = {Adv. Neural Inf. Process. Syst. (NeurIPS)},
  year      = {2020}
}

@inproceedings{Peebles_2023,
  author    = {William Peebles and Saining Xie},
  title     = {Scalable Diffusion Models with Transformers},
  booktitle = {Proc. IEEE/CVF Int. Conf. Comput. Vis. (ICCV)},
  year      = {2023},
  note      = {Best Paper Award}
}

@book{Birglen_2008,
  author    = {Lionel Birglen and Thierry Lalibert{\'e} and Cl{\'e}ment Gosselin},
  title     = {Underactuated Robotic Hands},
  series    = {Springer Tracts in Advanced Robotics},
  volume    = {40},
  publisher = {Springer},
  year      = {2008},
  isbn      = {978-3-540-77458-7}
}

@inproceedings{Zadaianchuk_2023,
  author    = {Andrii Zadaianchuk and Maximilian Seitzer and Georg Martius},
  title     = {Object-Centric Learning for Real-World Videos by Predicting Temporal Feature Similarities},
  booktitle = {Advances in Neural Information Processing Systems (NeurIPS)},
  year      = {2023}
}

@article{Simeoni_2025,
  author  = {Oriane Sim{\'e}oni and Huy V. Vo and Maximilian Seitzer and Federico Baldassarre and Maxime Oquab and others},
  title   = {{DINOv3}},
  journal = {arXiv preprint arXiv:2508.10104},
  year    = {2025}
}

@article{Balestriero_2025,
  author  = {Randall Balestriero and Yann LeCun},
  title   = {{LeJEPA}: Provable and Scalable Self-Supervised Learning Without the Heuristics},
  journal = {arXiv preprint arXiv:2511.08544},
  year    = {2025}
}

@article{Nam_2026,
  author  = {Heejeong Nam and Quentin Le Lidec and Lucas Maes and Yann LeCun and Randall Balestriero},
  title   = {Causal-{JEPA}: Learning World Models through Object-Level Latent Interventions},
  journal = {arXiv preprint arXiv:2602.11389},
  year    = {2026}
}

@article{Maes_2026_LeWM,
  author  = {Lucas Maes and Quentin Le Lidec and Damien Scieur and Yann LeCun and Randall Balestriero},
  title   = {{LeWorldModel}: Stable End-to-End Joint-Embedding Predictive Architecture from Pixels},
  journal = {arXiv preprint arXiv:2603.19312},
  year    = {2026}
}

@article{Wang_2026_AdaJEPA,
  author  = {Ying Wang and Oumayma Bounou and Yann LeCun and Mengye Ren},
  title   = {{AdaJEPA}: An Adaptive Latent World Model},
  journal = {arXiv preprint arXiv:2606.32026},
  year    = {2026}
}

@article{Hafner_2023,
  author  = {Danijar Hafner and Jurgis Pasukonis and Jimmy Ba and Timothy Lillicrap},
  title   = {Mastering Diverse Domains through World Models},
  journal = {arXiv preprint arXiv:2301.04104},
  year    = {2023}
}

@inproceedings{Hansen_2024,
  author    = {Nicklas Hansen and Hao Su and Xiaolong Wang},
  title     = {{TD-MPC2}: Scalable, Robust World Models for Continuous Control},
  booktitle = {International Conference on Learning Representations (ICLR)},
  year      = {2024}
}

@article{Kipf_2021_SAVi,
  author  = {Thomas Kipf and Gamaleldin F. Elsayed and Aravindh Mahendran and Austin Stone and Sara Sabour and Georg Heigold and Rico Jonschkowski and Alexey Dosovitskiy and Klaus Greff},
  title   = {Conditional Object-Centric Learning from Video},
  journal = {arXiv preprint arXiv:2111.12594},
  year    = {2021}
}

@inproceedings{Elsayed_2022_SAVipp,
  author    = {Gamaleldin F. Elsayed and Aravindh Mahendran and Sjoerd van Steenkiste and Klaus Greff and Michael C. Mozer and Thomas Kipf},
  title     = {{SAVi++}: Towards End-to-End Object-Centric Learning from Real-World Videos},
  booktitle = {Advances in Neural Information Processing Systems (NeurIPS)},
  year      = {2022}
}

@inproceedings{Seitzer_2023_DINOSAUR,
  author    = {Maximilian Seitzer and Max Horn and Andrii Zadaianchuk and Dominik Zietlow and Tianjun Xiao and Carl-Johann Simon-Gabriel and Tong He and Zheng Zhang and Bernhard Sch{\"o}lkopf and Thomas Brox and Francesco Locatello},
  title     = {Bridging the Gap to Real-World Object-Centric Learning},
  booktitle = {International Conference on Learning Representations (ICLR)},
  year      = {2023},
  note      = {{DINOSAUR}; arXiv:2209.14860}
}

@inproceedings{Wu_2023_SlotFormer,
  author    = {Ziyi Wu and Nikita Dvornik and Klaus Greff and Thomas Kipf and Animesh Garg},
  title     = {{SlotFormer}: Unsupervised Visual Dynamics Simulation with Object-Centric Models},
  booktitle = {International Conference on Learning Representations (ICLR)},
  year      = {2023}
}

@article{Chen_2020_ROOTS,
  author  = {Chang Chen and Fei Deng and Sungjin Ahn},
  title   = {{ROOTS}: Object-Centric Representation and Rendering of 3D Scenes},
  journal = {arXiv preprint arXiv:2006.06130},
  year    = {2020}
}

@article{Li_2021_MulMON,
  author  = {Nanbo Li and Cian Eastwood and Robert B. Fisher},
  title   = {Learning Object-Centric Representations of Multi-Object Scenes from Multiple Views},
  journal = {arXiv preprint arXiv:2111.07117},
  year    = {2021}
}

@inproceedings{Assran_2023_IJEPA,
  author    = {Mahmoud Assran and Quentin Duval and Ishan Misra and Piotr Bojanowski and Pascal Vincent and Michael Rabbat and Yann LeCun and Nicolas Ballas},
  title     = {Self-Supervised Learning from Images with a Joint-Embedding Predictive Architecture},
  booktitle = {Proc. IEEE/CVF Conf. Comput. Vis. Pattern Recognit. (CVPR)},
  year      = {2023}
}

@article{Assran_2025_VJEPA2,
  author  = {Mahmoud Assran and Adrien Bardes and David Fan and Quentin Garrido and Russell Howes and others},
  title   = {{V-JEPA 2}: Self-Supervised Video Models Enable Understanding, Prediction and Planning},
  journal = {arXiv preprint arXiv:2506.09985},
  year    = {2025}
}

@inproceedings{Hafner_2019_PlaNet,
  author    = {Danijar Hafner and Timothy Lillicrap and Ian Fischer and Ruben Villegas and David Ha and Honglak Lee and James Davidson},
  title     = {Learning Latent Dynamics for Planning from Pixels},
  booktitle = {International Conference on Machine Learning (ICML)},
  year      = {2019}
}

@inproceedings{Hafner_2020_Dreamer,
  author    = {Danijar Hafner and Timothy Lillicrap and Jimmy Ba and Mohammad Norouzi},
  title     = {Dream to Control: Learning Behaviors by Latent Imagination},
  booktitle = {International Conference on Learning Representations (ICLR)},
  year      = {2020}
}

@inproceedings{Finn_2017_VisualForesight,
  author    = {Chelsea Finn and Sergey Levine},
  title     = {Deep Visual Foresight for Planning Robot Motion},
  booktitle = {Proc. IEEE Int. Conf. Robot. Autom. (ICRA)},
  year      = {2017}
}

@article{Ebert_2018_VisualMPC,
  author  = {Frederik Ebert and Chelsea Finn and Sudeep Dasari and Annie Xie and Alex Lee and Sergey Levine},
  title   = {Visual Foresight: Model-Based Deep Reinforcement Learning for Vision-Based Robotic Control},
  journal = {arXiv preprint arXiv:1812.00568},
  year    = {2018}
}

@inproceedings{Manhaes_2016_UUVSim,
  author    = {Musa Morena Marcusso Manh{\~a}es and Sebastian A. Scherer and Martin Voss and Luiz Ricardo Douat and Thomas Rauschenbach},
  title     = {{UUV} Simulator: A Gazebo-Based Package for Underwater Intervention and Multi-Robot Simulation},
  booktitle = {OCEANS 2016 MTS/IEEE Monterey},
  year      = {2016},
  doi       = {10.1109/OCEANS.2016.7761080}
}

@article{Gazzaev_2026_AquaJEPA,
  author  = {Alan-Barsag Gazzaev and Alexey Gavrilov and Sergey Muravyov},
  title   = {{AquaJEPA}: Action-Conditioned Multimodal Predictive Representations for Underwater Robot Dynamics},
  journal = {arXiv preprint arXiv:2607.29393},
  year    = {2026},
  url     = {https://arxiv.org/abs/2607.29393}
}

@article{Cong_2021,
  author  = {Yang Cong and Changjun Gu and Tao Zhang and Yajun Gao},
  title   = {Underwater Robot Sensing Technology: A Survey},
  journal = {Fundamental Research},
  volume  = {1},
  number  = {3},
  pages   = {337--345},
  year    = {2021},
  doi     = {10.1016/j.fmre.2021.03.002}
}

@article{Huy_2023,
  author  = {Dinh Quang Huy and Nicholas Sadjoli and Abu Bakr Azam and Basman Elhadidi and Yiyu Cai and Gerald Seet},
  title   = {Object Perception in Underwater Environments: A Survey on Sensors and Sensing Methodologies},
  journal = {Ocean Engineering},
  volume  = {266},
  pages   = {113202},
  year    = {2023},
  doi     = {10.1016/j.oceaneng.2022.113202}
}

@article{Campos_2021_ORBSLAM3,
  author  = {Carlos Campos and Richard Elvira and Juan J. G{\'o}mez Rodr{\'i}guez and Jos{\'e} M. M. Montiel and Juan D. Tard{\'o}s},
  title   = {{ORB-SLAM3}: An Accurate Open-Source Library for Visual, Visual--Inertial, and Multimap {SLAM}},
  journal = {IEEE Transactions on Robotics},
  volume  = {37},
  number  = {6},
  pages   = {1874--1890},
  year    = {2021},
  doi     = {10.1109/TRO.2021.3075644}
}

@article{Zhang_2022_UVSSLAM,
  author  = {Nong Zhang and Shili Zhao and Dong An and Jincun Liu and He Wang and Yu Feng and Daoliang Li and Ran Zhao},
  title   = {Visual {SLAM} for Underwater Vehicles: A Survey},
  journal = {Computer Science Review},
  volume  = {46},
  pages   = {100510},
  year    = {2022},
  doi     = {10.1016/j.cosrev.2022.100510}
}

@inproceedings{Joshi_2019_VIOCompare,
  author    = {Bharat Joshi and Sharmin Rahman and Michail Kalaitzakis and Brennan Cain and James Johnson and Marios Xanthidis and Nare Karapetyan and Alan Hernandez and Alberto Quattrini Li and Nikolaos Vitzilaios and Ioannis Rekleitis},
  title     = {Experimental Comparison of Open Source Visual-Inertial-Based State Estimation Algorithms in the Underwater Domain},
  booktitle = {Proc. IEEE/RSJ Int. Conf. Intell. Robots Syst. (IROS)},
  pages     = {7227--7233},
  year      = {2019},
  doi       = {10.1109/IROS40897.2019.8968049}
}

@article{Rahman_2022_SVIn2,
  author  = {Sharmin Rahman and Alberto Quattrini Li and Ioannis Rekleitis},
  title   = {{SVIn2}: A Multi-Sensor Fusion-Based Underwater {SLAM} System},
  journal = {The International Journal of Robotics Research},
  volume  = {41},
  number  = {11--12},
  pages   = {1027--1047},
  year    = {2022},
  doi     = {10.1177/02783649221110259}
}

@article{Lyu_2023_StructuredLight,
  author  = {Nenqing Lyu and Haotian Yu and Jing Han and Dongliang Zheng},
  title   = {Structured Light-Based Underwater 3-D Reconstruction Techniques: A Comparative Study},
  journal = {Optics and Lasers in Engineering},
  volume  = {161},
  pages   = {107344},
  year    = {2023},
  doi     = {10.1016/j.optlaseng.2022.107344}
}

@article{Ou_2023_WaterMBSL,
  author  = {Yaming Ou and Junfeng Fan and Chao Zhou and Long Cheng and Min Tan},
  title   = {{Water-MBSL}: Underwater Movable Binocular Structured Light-Based High-Precision Dense Reconstruction Framework},
  journal = {IEEE Transactions on Industrial Informatics},
  year    = {2023},
  doi     = {10.1109/TII.2023.3342899}
}

@article{DulacArnold_2019,
  author  = {Gabriel Dulac-Arnold and Daniel Mankowitz and Todd Hester},
  title   = {Challenges of Real-World Reinforcement Learning},
  journal = {arXiv preprint arXiv:1904.12901},
  year    = {2019}
}

@inproceedings{Chua_2018_PETS,
  author    = {Kurtland Chua and Roberto Calandra and Rowan McAllister and Sergey Levine},
  title     = {Deep Reinforcement Learning in a Handful of Trials using Probabilistic Dynamics Models},
  booktitle = {Adv. Neural Inf. Process. Syst. (NeurIPS)},
  year      = {2018}
}

@article{Bicchi_2000,
  author  = {Antonio Bicchi},
  title   = {Hands for Dexterous Manipulation and Robust Grasping: A Difficult Road Toward Simplicity},
  journal = {IEEE Transactions on Robotics and Automation},
  volume  = {16},
  number  = {6},
  pages   = {652--662},
  year    = {2000},
  doi     = {10.1109/70.897777}
}

@book{Featherstone_2008,
  author    = {Roy Featherstone},
  title     = {Rigid Body Dynamics Algorithms},
  publisher = {Springer},
  year      = {2008},
  isbn      = {978-0-387-74314-1}
}

@inproceedings{Todorov_2012_MuJoCo,
  author    = {Emanuel Todorov and Tom Erez and Yuval Tassa},
  title     = {{MuJoCo}: A Physics Engine for Model-Based Control},
  booktitle = {Proc. IEEE/RSJ Int. Conf. Intell. Robots Syst. (IROS)},
  pages     = {5026--5033},
  year      = {2012},
  doi       = {10.1109/IROS.2012.6386109}
}

@article{Yuan_2017_GelSight,
  author  = {Wenzhen Yuan and Siyuan Dong and Edward Adelson},
  title   = {{GelSight}: High-Resolution Robot Tactile Sensors for Estimating Geometry and Force},
  journal = {Soft Robotics},
  volume  = {4},
  number  = {3},
  pages   = {253--263},
  year    = {2017},
  doi     = {10.1089/soro.2016.0053}
}

@article{Lambeta_2020_DIGIT,
  author  = {Mike Lambeta and Po-Wei Chou and Stephen Tian and Brian H. Yang and Benjamin Maloon and Victoria Rose Most and Dave Stroud and Raymond Santos and Ahmad Byagowi and Gregg Kammerer and Dinesh Jayaraman and Roberto Calandra},
  title   = {{DIGIT}: A Novel Design for a Low-Cost Compact High-Resolution Tactile Sensor with Application to In-Hand Manipulation},
  journal = {IEEE Robotics and Automation Letters},
  volume  = {5},
  number  = {3},
  pages   = {3838--3845},
  year    = {2020},
  doi     = {10.1109/LRA.2020.2977257}
}

@inproceedings{Wu_2023_DayDreamer,
  author    = {Philipp Wu and Alejandro Escontrela and Danijar Hafner and Pieter Abbeel and Ken Goldberg},
  title     = {{DayDreamer}: World Models for Physical Robot Learning},
  booktitle = {Proc. Conf. Robot Learning (CoRL)},
  series    = {Proceedings of Machine Learning Research},
  volume    = {205},
  pages     = {2226--2240},
  year      = {2023}
}

@article{Yang_2025_Salvage,
  author  = {Yuncong Yang and Chunwen Zhang and Feng Wu and Jinlong Li and Xuyang Wang},
  title   = {Rapid Salvage of Deep-Sea Underactuated Manipulators: Real-Time Solution of Contact Forces and Impact Assessment of Disturbance Forces},
  journal = {Robot},
  year    = {2025},
  doi     = {10.13973/j.cnki.robot.240001},
  note    = {(in Chinese)}
}

@misc{Rao_2026_SkyJEPA,
  author = {Pratyaksh Rao and Wancong Zhang and Randall Balestriero and Yann LeCun and Giuseppe Loianno},
  title  = {{SkyJEPA}: Learning Long-Horizon World Models for Zero-Shot Sim-to-Real Control of Quadrotors},
  note   = {arXiv:2606.23444},
  year   = {2026}
}

@misc{Cheng_2026_SRWm,
  author = {Juntao Cheng and Jingkai Wang and Yijun Shen and Xiansheng Chen and Zhiwei Yu},
  title  = {Beyond Instance Slots: Semantically Rich World Models for Physical Interaction Planning},
  note   = {arXiv:2608.22294},
  year   = {2026}
}

@misc{Khan_2026_DepthReg,
  author = {Usman M. Khan},
  title  = {Depth-Regularized {JEPA} World Models Learn More Transferable Representations from Real Outdoor Robot Data},
  note   = {arXiv:2607.16314},
  year   = {2026}
}

@article{Zhang_2026_PICWM,
  author  = {Bolun Zhang and Canjun Yang},
  title   = {Physics-Informed Compact World Models for Predictive Control in {AUV} Terminal Docking under Communication Constraints},
  journal = {International Journal of Intelligent Robotics and Applications},
  year    = {2026},
  doi     = {10.1007/s41315-026-00530-1}
}

@inproceedings{Weinzaepfel_2022_CroCo,
  author    = {Weinzaepfel, Philippe and Leroy, Vincent and Lucas, Thomas and Br{\'e}gier, Romain and Cabon, Yohann and Arora, Vaibhav and Antsfeld, Leonid and Chidlovskii, Boris and Csurka, Gabriela and Revaud, J{\'e}r{\^o}me},
  title     = {{CroCo}: Self-Supervised Pre-training for {3D} Vision Tasks by Cross-View Completion},
  booktitle = {Advances in Neural Information Processing Systems (NeurIPS)},
  year      = {2022}
}

@inproceedings{Seo_2023_MVMWM,
  author    = {Seo, Younggyo and Kim, Junsu and James, Stephen and Lee, Kimin and Shin, Jinwoo and Abbeel, Pieter},
  title     = {Multi-View Masked World Models for Visual Robotic Manipulation},
  booktitle = {Proceedings of the 40th International Conference on Machine Learning (ICML)},
  year      = {2023}
}

@misc{Zhang_2026_ThinkJEPA,
  author = {Haichao Zhang and Yijiang Li and Shwai He and Tushar Nagarajan and Mingfei Chen and Jianglin Lu and Ang Li and Yun Fu},
  title  = {{ThinkJEPA}: Empowering Latent World Models with Large Vision-Language Reasoning Model},
  note   = {arXiv:2603.22281},
  year   = {2026}
}

@inproceedings{Garrido_2023_RankMe,
  author    = {Garrido, Quentin and Balestriero, Randall and Najman, Laurent and LeCun, Yann},
  title     = {{RankMe}: Assessing the Downstream Performance of Pretrained Self-Supervised Representations by Their Rank},
  booktitle = {Proceedings of the 40th International Conference on Machine Learning (ICML)},
  year      = {2023}
}

@misc{Wen_2026_JEPAx,
  author = {Kehan Wen and Ziming Li and Siyuan Luo and Fan Shi},
  title  = {{JEPA-x}: Cross-Predictive Physics Grounding for Forecastable Latent Dynamics},
  note   = {arXiv:2608.24044},
  year   = {2026}
}

@misc{Hang_2026_Habit,
  author = {Jinting Hang and Zhenhui Cai},
  title  = {Identifying Habit, Physics, and Nuisance in Robot World Models},
  note   = {arXiv:2609.09210},
  year   = {2026}
}

@misc{Wang_2026_DMA,
  author = {Jiawei Wang and Ke Rui and Yushen Zuo and Yichun Feng and Minglei Li},
  title  = {Decision-Metric Alignment in Latent World Models: Diagnostics and Action-Conditioned Objectives for {MPC} Planning},
  note   = {arXiv:2608.18746},
  year   = {2026}
}

@misc{Aljalbout_2026_RealityGap,
  author = {Elie Aljalbout and others},
  title  = {The Reality Gap in Robotics: Challenges, Solutions, and Best Practices},
  note   = {arXiv:2510.20808},
  year   = {2026}
}

\end{document}